\documentclass[preprint]{elsarticle}

\makeatletter
\def\ps@pprintTitle{%
	\let\@oddhead\@empty
	\let\@evenhead\@empty
	\def\@oddfoot{\footnotesize\centerline{\thepage}\hfil
		\today}
	\let\@evenfoot\@oddfoot
}
\makeatother

\usepackage{url}
\usepackage{tabularx}
\usepackage{changepage}

\usepackage{algorithm}
\usepackage{algpseudocode}
\usepackage{algorithmicx}
\usepackage{hyperref}

\usepackage{multirow}
\usepackage{rotating}

\usepackage{booktabs}

\usepackage{mwe}

\usepackage{xcolor}

\algnewcommand{\Input}[1]{%
	\Statex \textbf{Input:}
	\Statex \hspace*{\algorithmicindent}\parbox[t]{.8\linewidth}{\raggedright #1}
}

\algnewcommand{\Output}[1]{%
	\Statex \textbf{Output:}
	\Statex \hspace*{\algorithmicindent}\parbox[t]{.8\linewidth}{\raggedright #1}
}

\algnewcommand{\IIf}[1]{\State\algorithmicif\ #1\ \algorithmicthen}
\algnewcommand{\EndIIf}{\unskip\ \algorithmicend\ \algorithmicif}

\usepackage{subcaption}

\renewcommand{\thesubfigure}{.\arabic{subfigure}}

\usepackage{makecell}
\usepackage{rotating}

\usepackage{lineno}

\begin{document}	
	\begin{frontmatter}
		\title{Multivariable times series classification through an interpretable representation}
		\author[DECSAI]{Francisco J.~Bald\'an\corref{cor1}}
		\ead{fjbaldan@decsai.ugr.es}
		\author[DECSAI]{Jos\'e M.~Ben\'itez}
		\ead{J.M.Benitez@decsai.ugr.es}
		
		\cortext[cor1]{Corresponding author}
		
		\address[DECSAI]{Department of Computer Science and Artificial Intelligence, University of Granada, DICITS, iMUDS, DaSCI, 18071 Granada, Spain}

		\begin{abstract}
			Multivariate time series classification is a task with increasing importance due to the proliferation of new problems in various fields (economy, health, energy, transport, crops, etc.) where a large number of information sources are available. Direct extrapolation of methods that traditionally worked in univariate environments cannot frequently be applied to obtain the best results in multivariate problems. This is mainly due to the inability of these methods to capture the relationships between the different variables that conform a multivariate time series. The multivariate proposals published to date offer competitive results but are hard to interpret. In this paper we propose a time series classification method that considers an alternative representation of time series through a set of descriptive features taking into account the relationships between the different variables of a multivariate time series. We have applied traditional classification algorithms obtaining interpretable and competitive results.
			
		\end{abstract}
	
	\end{frontmatter}
\section{Introduction}

Nowadays, large amounts of data are generated. Everything is increasingly interconnected, more and more sensors are included in everything around us, and these monitor the behavior of any event of interest over time. These sensors generate lots of data in the form of multivariate time series (MTS). A key task in the analysis and mining of these data is multivariate time series classification (MTSC), which aims to give an accurate response to a large number of problems: e.g. from detecting when a patient is sick or has an anomaly in his heart behavior~\cite{maharaj2014discriminant}, or if a driver is in optimal condition to drive~\cite{li2015drunk}, the recognition of human activities~\cite{seto2015multivariate} or how to adapt energy production based on particular circumstances~\cite{khan2018load}.

The field of MTSC can be divided into two main types of work. Firstly, applied works that seek to obtain a better solution for a given problem, offering ad-hoc proposals considering the peculiarities of the treated problem~\cite{chaovalitwongse2011pattern}\cite{mcgovern2011identifying}\cite{klockl2008multivariate}. Secondly, proposals that deal with MTS in a general way but taking into account possible interrelations between the different variables available~\cite{orsenigo2010combining}\cite{antonucci2015robust}\cite{gorecki2015multivariate}\cite{yu2019fast}. The proposals in the later group are usually based on strong theoretical foundations. A relatively large number of proposals for MTSC can be found in the literature~\cite{schafer2017multivariate}\cite{baydogan2015learning}\cite{karim2019multivariate}\cite{fauvel2020local}. Most of them are guided towards obtaining increasing levels of accuracy. However, eXplainable Artificial Intelligence (XAI)~\cite{sosilovic2018explainable} is a topic enjoying a growing level of interest. Its goal is to build accurate intelligent system for complex tasks, but also paying special attention to their interpretability. The built systems or the way they make decisions are required to be easy to understand for human beings. Thus high accuracy is no longer the only objective, interpretability receives higher attention. This also applies to solutions for classification problems. In the field of MTSC, there are few proposals that pay attention to the interpretability of results~\cite{grabocka2016fast}. Given the complexity of the problem, most proposals are focused on obtaining the best results in terms of accuracy. Even the proposals based on shapelets~\cite{ye2009time}\cite{baldan2019distributed}, which are interpretable from their univariate origins, have chosen to use the Transformed Shapelets in multivariate environments~\cite{bostrom2017shapelet} or proposals that are even less interpretable~\cite{karimi2018scalable}, giving priority to accuracy results over interpretability. One possible way to pave de path towards easier to understand solutions to MTSC is expressing time series in different domains. Perhaps in terms of descriptive features instead of the raw time domain values.

In this paper, we present a new MTSC approach based on the representation of time series through a set of features and measures. This approach allows transforming the original MTSC problem into a traditional classification problem, enabling to apply the whole set of the traditional classification algorithms. While focused on raising the interpretability of the classification results, the approach allows to obtain competitive accuracy results with respect to the main techniques of the state-of-the-art.

The remainder of this paper is organized as follows: Section~\ref{background} introduces the state of the art in MTSC. Section~\ref{CMF_proposal} describes in depth our proposal. Section~\ref{Empirical Study} shows the experimental study conducted, the results obtained, and the interpretability of them. Finally, Section~\ref{conclusion} concludes the paper.

\section{Related work}
\label{background}
In the field of MTSC, proposals from methods that have demonstrated good behaviour in univariate cases predominate. Some of the first proposals for MTSC were multivariate extensions of the distance-based algorithm 1NN-DTW~\cite{vlachos2003indexing}\cite{keogh2001derivative}, given its simplicity and good results in univariate environments. These proposals are a good starting point but they carry the limitations they had in univariate environments: high computational complexity and low interpretability, since they only indicate how much the instances are similar to each other. To these limitations we must add that in a multivariable environment the first proposals of 1NN-DTW processed each variable of each time series independently, so they were not able to extract information from the relationship between the different variables that make up each multivariate time series. With this in mind, we can say that multiple proposals for a multivariate DTW have been made, such as dependent (DTW$_{D}$) and independent (DTW$_{I}$) warping, both having the same performance~\cite{shokoohi2017generalizing}. Other proposals such as Mahalanobis Distance-based Dynamic Time Warping measure (MDDTW)~\cite{mei2015learning} seek to give a general answer to this problem. MDDTW is able to precisely calculate the relationship between the different variables that compose an MTS, this together with the alignment obtained by DTW allows to obtain very competitive results.

The feature-based approach has multiple proposals, giving special importance to the extraction of additional information and to the speed of processing, especially when compared to similarity-based techniques. In this field we can differentiate between proposals based on shapelets and bag-of-words. In the field of shapelets, Generalized Random Shapelets Forests (gRSF)~\cite{karlsson2016generalized} is considered the state-of-the-art, obtaining better results than its direct competitor, Ultra Fast Shapelets (UFS)~\cite{wistuba2015ultra}. gRSF is based on the creation of a set of shapelet-based decision trees from a random extraction of the shapelets. In the field of bag-of-words, Word ExtrAction for time Series cLassification plus Multivariate Unsupervised Symbols and dErivatives (WEASEL+MUSE)~\cite{schafer2017multivariate} is considered the state-of-the-art, as it obtains the best results against its direct competitors: Learned Pattern Similarity (LPS)~\cite{baydogan2016time}, Autoregressive forests for multivariate time series modelling (mv-ARF)~\cite{tuncel2018autoregressive}, Symbolic representation for Multivariate Time Series classification (SMTS)~\cite{baydogan2015learning}, and gRSF. All of them have been tested on one of the first reference MTS database collected from~\cite{baydogan2020Datasets}, with a total of 20 MTSC datasets. WEASEL+MUSE extracts a vector of features by applying a sliding-window to each variable of the MTSC and filtering out non-discriminative features, finally a classifier analyses these data.

In the field of deep learning, the extension of the Long Short Term Memory Fully Convolutional Network (LSTM-FCN) and Attention LSTM-FCN (ALSTM-FCN)~\cite{karim2019multivariate} to a multivariate environment, including a squeeze-and-excitation block in the fully convolutional block that improves accuracy. This proposal improved the WEASEL+MUSE results over the original database of 20 datasets~\cite{baydogan2020Datasets} extended with 10 datasets from the UC Irvine Machine Learning Repository (UCI)~\cite{dua2019machine} and 6 datasets used by Pei et al.~\cite{pei2017multivariate}

A new proposal has recently emerged, Local Cascade Ensemble for Multivariate Data Classification (LCE) and its extension for Multivariate Time Series (LCEM)~\cite{fauvel2020local}. LCE and LCEM are a hybrid ensemble method with 2 major objectives. The first one is to handle the bias-variance tradeoff by an explicit boosting-bagging approach. The second one is to individualize classifier errors on different parts of the training data by an implicit divide-and-conquer approach. This proposal is outlined as the new state-of-the-art in MTSC by obtaining better results than the previous state-of-the-art MLSTM-FCN and WEASEL+MUSE, on the University of East Anglia (UEA) repository~\cite{bagnall2018uea}, a new repository for MTSC composed of 30 datasets that is becoming increasingly important.

In contrast to state-of-the-art methods, we propose a method that obtains essential features of each variable and each MTS and applies a transformation to the MTS dataset obtaining a traditional classification problem based on attributes. All traditional classification algorithms can be applied to this new dataset, and depending on the applied algorithms, interpretable results can be obtained to explain the problem or results of higher accuracy.

\section{Multivariable times series classification through an interpretable representation}
\label{CMF_proposal}
In this work we propose a method that allows the calculation of complexity measures to be applied to MTSC problems. Our proposal, namely Complexity Measures and Features for Multivariate Time Series (CMFMTS), is based on the idea that a time series can be faithfully represented with a set of complexity measures and descriptive features~\cite{baldan2020complexity}. Furthermore, these features preserve most of the information content of the series to such an extend that they can be used to classify the series. 

The following is an example of the calculation of some features on an MTS with three variables. In Table~\ref{somecm}, we show some features highly related to the nature of the time series and its range of possible values. In Figure~\ref{fig:tsexample}, we show a simple example of the feature extraction used and its interpretability. In the first place, we can see how variables 1 and 2 are similar, so we can expect values of the features also similar to each other. This is reflected in the values of kurtosis and skewness. If variables 1 and 2 have similar values their probability distribution will be similar and therefore their values of kurtosis and skewness. We can appreciate a significant difference concerning variable 3. In the three variables, we can see the existence of a single trend, for this reason, the trend values are close to 1 in all cases. The oscillations present in the variables 1 and 2 seem more typical of seasonal patterns that do not affect the trend of the time series. To evaluate the Chao-Shen shannon entropy (shannon\_entropy\_cs) we have to appreciate the evolution of variables 1, 2, and 3. Variables 1 and 2 show a certain pattern, while variable 3 shows a long period without changes with a final reduction of the value never seen before. For this reason, it obtains a higher value in shannon entropy very far from the one obtained by variables 1 and 2. We also analyze the values of curvature and linearity. Given the evolution and shape of the three variables and the perceptible linear relationship between the current values of variables 1 and 2 with their corresponding past values, it is logical to expect positive and similar values of curvature and linearity for variables 1 and 2. On the other hand, variable 3 does not show these forms or a linear relationship between its present and past values, so it obtains negative values that are far from curvature and linearity concerning what is obtained by variables 1 and 2.

\begin{table}[!ht]
	\caption{Example of some features used.}
	\label{somecm} 
	\begin{adjustwidth}{-1.7cm}{}
		\begin{tabularx}{1.25\textwidth}{|llXl|}
			\toprule
			Char. & Name                    & Description                                                                   & Range                \\
			\midrule
			$F_{1}$  & curvature          & Calculated based on the coefficients of an orthogonal quadratic regression        &       $(-\infty,\infty)$                  \\
			$F_{2}$             & kurtosis                & The ``tailedness" of the probability distribution                   & $(-\infty,\infty)$                      \\
			$F_{3}$  & linearity          & Calculated based on the coefficients of an orthogonal quadratic regression      &         $(-\infty,\infty)$                   \\
			$F_{4}$              & shannon\_entropy\_cs    & Chao-Shen entropy estimator                                                   & $[0,\infty)$                    \\
			$F_{5}$             & skewness                & Asymmetry of the probability distribution                          & $(-\infty,\infty)$                     \\
			$F_{6}$  & trend              & Strength of trend           &               $[0,1]$           \\
			\bottomrule
		\end{tabularx}
	\end{adjustwidth}
\end{table}

\begin{figure}[!ht]
	\includegraphics[width=\textwidth]{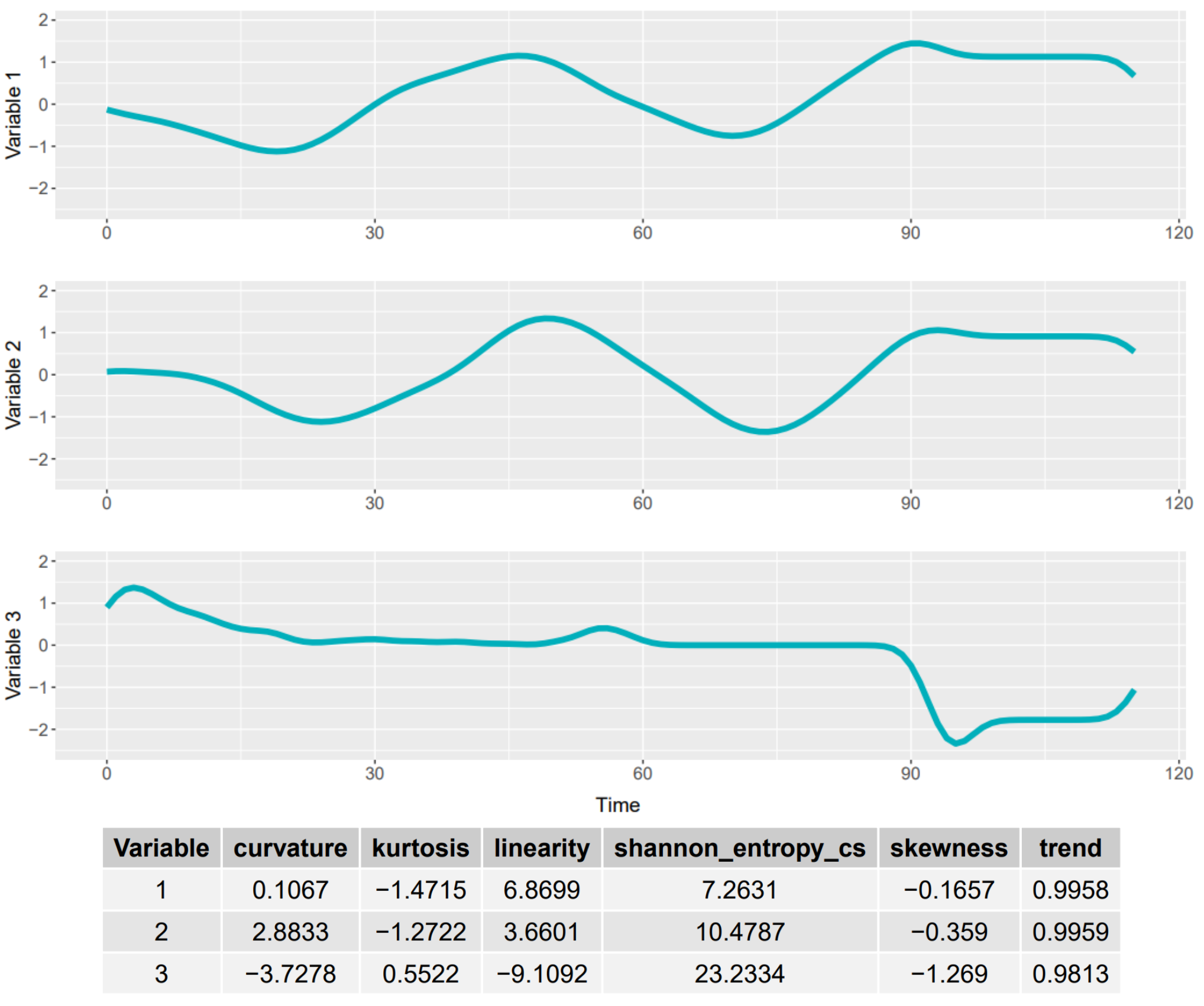}
	\caption{Example of feature extraction from an MTS with 3 variables.}
	\label{fig:tsexample}
\end{figure} 

\begin{figure}[!ht]
	\includegraphics[width=\textwidth]{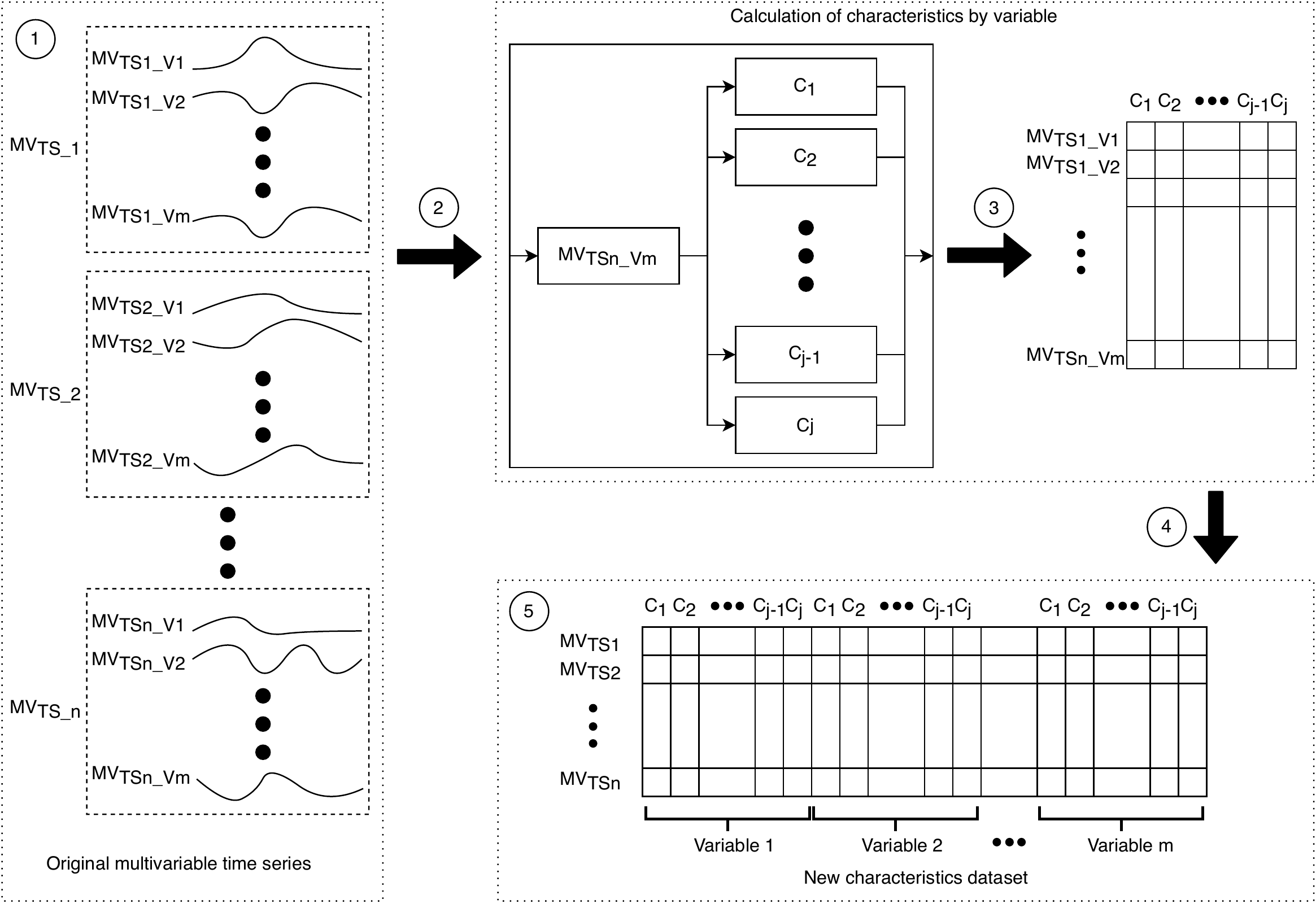}
	\caption{Features calculation workflow.}
	\label{fig:proposaldiagram}
	\setcounter{figure}{0} 
	{\phantomsubcaption\label{sub1}}
	{\phantomsubcaption\label{sub2}}
	{\phantomsubcaption\label{sub3}}
	{\phantomsubcaption\label{sub4}}
	{\phantomsubcaption\label{sub5}}
\end{figure}

Figure~\ref{fig:proposaldiagram} shows the workflow of our proposal. First, a set of $n$ multivariate time series is assumed, each consisting of $m$ variables (Figure~\ref{sub1}). Individually, each one of the variables that compose each time series is processed, obtaining the $j$ features for each variable (Figure~\ref{sub2}). A dataset is obtained with $n \times m$ rows and $j$ columns, where each row is composed of the set of features processed on each time series that compose each MTS (Figure~\ref{sub3}). Finally, this dataset is processed by placing all the variables of the same MTS in the same row (Figure~\ref{sub4}). In this way, a new (transformed) dataset is obtained where all the features of all the variables that compose the same MTS are placed in the same row, forming part of the same instance. This enables the search for patterns and relationships of interest among the different variables that compose the same MTS (Figure~\ref{sub5}).

Although the feature calculation based approach can be applied to all types of automatic learning problems as well as supervised, unsupervised, semi-supervised learning, etc. In the supervised case, simple and fast comparisons can be made with respect to the main state-of-the-art algorithms.
Due to the great variety of the processed time series, it is possible that undesired values are obtained for some of the proposed features. For example, to calculate the autocorrelation coefficient function (ACF)~\cite{baldan2020complexity} concerning the values delayed 10 instants of time it is necessary that our time series has a minimum length of 11, otherwise, we will obtain an Not Available (NA). Time series with a single value is another problematic case since features like kurtosis and skewness are not defined for these cases and would return Not a Number (NaN) values. Time series containing NA generate problems internally in some of the features used (acf, kurtosis, skewness, shannon\_entropy\_cs, etc.) returning NA values in those features. Finally, there are features that can obtain values in the range $(-\infty,\infty)$. Extreme values close to the limits are considered as undesirable since they generate several problems in the different algorithms applied later. To deal with these cases, we have specified a preprocessing stage, following the calculation of the features and their correct ordering, which solves the possible inconveniences generated by these cases. The whole process is depicted in Algorithm~\ref{Algorithm1}.

\begin{algorithm}[!ht] 
	\floatname{algorithm}{Algorithm}
	\caption{Preprocessing procedure}
	\label{Algorithm1}
	\small
	\begin{algorithmic}[1]
		\Input{
			\textit{$train$}: train dataframe with (Ts\_id, Ts\_dimId, Ts\_class, Ts\_values)\\
			\textit{$test$}: test dataframe with (Ts\_id, Ts\_dimId, Ts\_class, Ts\_values) \\
			\textit{$models $}: list of models to be processed \\
		}
		\Output{$output\_data$: a triplet that contains the fitted models, the vectors with the predicted labels and the accuracies obtained \\
			$mvf\_train$: features train dataframe \\
			$mvf\_test$: features test dataframe  \\
		}
		\State mvf\_train, mvf\_test $\gets$ calc\_mvcmfts((train, test), all)

		\For {each value in (mvf\_train, mvf\_test)}
		\IIf {(is.na(value) $\|$ is.nan(value) $\|$ is.infinite(value))} value $\gets$ NA \EndIIf
		\EndFor
		
		\For {each column in mvf\_train}
		\If {sum(!is.na(colum.values)) $==$ 0)} 
		\State mvf\_train.delete(column)
		\State mvf\_test.delete(column)
		\EndIf
		\EndFor
		
		\For {each column in (mvf\_train, mvf\_test)}
		\For {each value in column}
		\IIf {is.na(value)} value $\gets$ mean(column)
		\EndIIf
		
		\EndFor
		\EndFor
		
		\For {each column in mvf\_train}
		\If {(length(unique(column)) $<=$ 1)} 
		\State mvf\_train.delete(column.index) 
		\State mvf\_test.delete(column.index) 
		\EndIf
		\EndFor
		
		\State output\_data $\gets$ NULL
		\For {each model in models}
		\State  fit $\gets$ train.model(mvf\_train, train.Ts\_class)
		\State	pred $\gets$ fit.predict(mvf\_test)
		\State 	acc $\gets$ accuracy(pred, test.Ts\_class)
		\State  output$\_$data.add(fit, pred, acc)
		\EndFor
		
		\State \textbf{return} (output\_data, mvf\_train, mvf\_test)
	\end{algorithmic}
\end{algorithm}

The starting point is the calculation of the proposed features in the training and test sets (Line 1). For any of the cases mentioned above in which an undesired value has been obtained, these values are unified under a single NA identifier (Lines 2-4). We check on the training set if any column lacks interest because it is full of undesired values. If so, this column is removed from both the training set and the test set (Lines 5-10). In order to simplify the treatment of missing values, we have chosen to impute these values with the average of their respective column (Lines 11-15). There are better imputation techniques, but we do not address that task in this paper and the considered one has proved to ve effective enough. To avoid the use of variables without information, we analyzed the training set looking for variables with a single value. If any variable with this condition is found, it is eliminated from both the training set and the test set (Lines 16-21). Finally, each of the specified models is processed, obtaining the desired model fit, its prediction on the test set and the accuracy achieved (Lines 23-28). These data are returned to the user, together with the datasets transformed to the features of our proposal (Line 29).

Finally, once we have explained our proposal and its application in a real environment, we can list the main advantages offered by this approach:

\begin{itemize}
	\item Allows the use of the application of any vector-based classification method, since after the applied transformation, we obtain a traditional dataset where each instance is represented by its corresponding attributes (features).
	\item Allows the use of machine learning methods based on different paradigms: supervised, semi-supervised, self-supervised, unsupervised, etc., since it obtains a vector-based dataset, where each instance is composed of different attributes.
	\item Handles easily datasets of time series with varying lengths, as it processes each time series individually.
	\item Decisions made can be easily understood by human experts, since the features used explain the behavior of the time series. In addition, the represented concepts by the selected features are interpretable by the users.
\end{itemize}

\section{Empirical Study}
\label{Empirical Study}
To assess the performance in classification tasks of the proposal a thorough empirical study has been designed and carried out. We start by explaining the experimental design (Section~\ref{Experimental Design}), followed by the results obtained (Section~\ref{Results}). Finally, we analyze the interpretability of the models obtained (Section~\ref{Interpretability}).

\subsection{Experimental Design}
\label{Experimental Design}
We describe the performance measures used to evaluate our proposal (Section~\ref{Performance measures}), followed by the datasets used (Section~\ref{Datasets}) and the machine learning models selected (Section~\ref{Models}). Finally, we describe the hardware and software used in the development of our proposal (Section~\ref{Hardware and Software}).

\subsubsection{Performance measures}
\label{Performance measures}
Since the datasets come from very different fields, we have opted for a ranking performance measure. We have selected the average rank as a comparative method, from the calculation of the accuracy on the original training and test sets. The accuracy has been calculated as the number of instances correctly classified divided the total number of instances of the test set. Also, we have included the Win/Loss/Tie ratio to quantify the number of cases in which each model and approach wins, loses, or ties concerning the best case. Since the range of possible results is wide, we have opted to include a Critical Difference diagram (CD)~\cite{demvsar2006statistical}. CD shows the results of statistical comparison between all models in pairs based on average ranks. Models that are connected by a bold line do not have a statistically significant difference, for a particular confidence level. In our case, we have set an $\alpha$ of 0.05, for a 95\% confidence level. The average rank and the CD were obtained using the R~\textit{scmamp} package

\subsubsection{Datasets}
\label{Datasets}
To evaluate the performance of our proposal on problems of all kinds, we have selected the main repository of MTSC problems, the UEA multivariate time series classification archive. In Table~\ref{UEAInfo}, we show the characteristics of the 30 datasets of the UEA repository: number of instances of the training and test sets, length of the time series, number of variables of each MTS and number of classes. Some of these datasets are composed by time series of different length, so the repository chose to padding with NA values. In our case we have removed those values. We have processed the values of the time series that contain information without affecting the original values of each time series.

\begin{table}[!ht]
	\centering
	\caption{Datasets information from the UEA repository}
	\label{UEAInfo}
	\resizebox{1\textwidth}{!}{
	\begin{tabular}{|l|l|l|l|l|l|}
		\toprule
		Dataset                   & Train & Test  & Length & Dims & Class \\
		\midrule
		ArticularyWordRecognition & 275   & 300   & 144    & 9    & 25    \\
		AtrialFibrillation        & 15    & 15    & 640    & 2    & 3     \\
		BasicMotions              & 40    & 40    & 100    & 6    & 4     \\
		CharacterTrajectories     & 1422  & 1436  & 60-182 & 3    & 20    \\
		Cricket                   & 108   & 72    & 1197   & 6    & 12    \\
		DuckDuckGeese             & 50    & 50    & 270    & 1345 & 5     \\
		EigenWorms                & 128   & 131   & 17984  & 6    & 5     \\
		Epilepsy                  & 137   & 138   & 206    & 3    & 4     \\
		EthanolConcentration      & 261   & 263   & 1751   & 3    & 4     \\
		ERing                     & 30    & 270   & 65     & 4    & 6     \\
		FaceDetection             & 5890  & 3524  & 62     & 144  & 2     \\
		FingerMovements           & 316   & 100   & 50     & 28   & 2     \\
		HandMovementDirection     & 160   & 74    & 400    & 10   & 4     \\
		Handwriting               & 150   & 850   & 152    & 3    & 26    \\
		Heartbeat                 & 204   & 205   & 405    & 61   & 2     \\
		InsectWingbeat            & 25000 & 25000 & 2-22   & 200  & 10    \\
		JapaneseVowels            & 270   & 370   & 7-29   & 12   & 9     \\
		Libras                    & 180   & 180   & 45     & 2    & 15    \\
		LSST                      & 2459  & 2466  & 36     & 6    & 14    \\
		MotorImagery              & 278   & 100   & 3000   & 64   & 2     \\
		NATOPS                    & 180   & 180   & 51     & 24   & 6     \\
		PenDigits                 & 7494  & 3498  & 8      & 2    & 10    \\
		PEMS-SF                   & 267   & 173   & 144    & 963  & 7     \\
		PhonemeSpectra            & 3315  & 3353  & 217    & 11   & 39    \\
		RacketSports              & 151   & 152   & 30     & 6    & 4     \\
		SelfRegulationSCP1        & 268   & 293   & 896    & 6    & 2     \\
		SelfRegulationSCP2        & 200   & 180   & 1152   & 7    & 2     \\
		SpokenArabicDigits        & 6599  & 2199  & 4-93   & 13   & 10    \\
		StandWalkJump             & 12    & 15    & 2500   & 4    & 3     \\
		UWaveGestureLibrary       & 120   & 320   & 315    & 3    & 8    \\
		\bottomrule
	\end{tabular}
}
\end{table}

\subsubsection{Models}
\label{Models}
For our proposal, we have selected a set of traditional models with two main approaches: to obtain interpretable results and to obtain the best classification results by sacrificing interpretability~\cite{baldan2020complexity}. These models are C5.0 with boosting (C5.0B), RandomForest (RF), Support Vector Machine (SVM), and 1-Nearest Neighbors with Euclidean Distance (1NN-ED). For this last model we have applied a normalization between [0, 1]. This set of models will be applied to the set of time series features obtained by our proposal. The final models of our proposal are obtained from the union of the transformed datasets with the four models previously commented. These proposals are: CMFMTS+C5B, CMFMTS+RF, CMFMTS+SVM, and CMFMTS+1NN-ED. We have simplified the CMFMTS nomenclature by CMFM due to space limitations in later tables. On the other hand, we have selected the main state-of-the-art MTSC models:

\begin{itemize}
	\item 1-Nearest Neighbor classifier with Euclidean distance (1NN-ED), with and without normalization.
	\item 1-Nearest Neighbor classifier based on multi-dimensional points (DTW-1NN-D)~\cite{shokoohi2017generalizing}, with and without normalization.
	\item 1-Nearest Neighbor classifier based on the sum of DTW distance for each dimension (DTW-1NN-I)~\cite{shokoohi2017generalizing}, with and without normalization.
	\item Multivariate LSTM Fully Convolutional Networks for Time Series Classification (MLSTM-FCN)~\cite{karim2019multivariate} with the settings specified by their authors: 128-256-128 filters, 250 training epochs, a dropout of 0.8 and a batchsize of 128.
	\item Word ExtrAction for time SEries cLassification plus Multivariate Unsupervised Symbols and dErivatives (WEASEL+MUSE)~\cite{schafer2017multivariate} with the settings specified by their authors: SFA word lengths l in [2,4,6], windows length in [4:max(MTSlength)], chi=2, bias=1, p=0.1, c=5 and a solver equals to L2R LR DUAL.
	\item Local Cascade Ensemble for Multivariate data classification (LCEM)~\cite{fauvel2020local}, optimized hyperparameters for each dataset (Windows (\%), Trees, and Depth). The results have been obtained from the published work of the authors.
	\item Random Forest for Multivariate (RFM) algorithm, from the sklearn library, applied to the transformation proposed in the LCEM paper~\cite{fauvel2020local}.
	\item Extreme Gradient Boosting for multivariate (XGBM), Extreme Gradient Boosting algorithm, from the xgboost library, applied to the transformation proposed in the LCEM paper~\cite{fauvel2020local}.
	
\end{itemize}

 	The results of the algorithms mentioned above have been obtained from~\cite{fauvel2020local}.

\subsubsection{Hardware and Software}
\label{Hardware and Software}
The experimentation carried out in this work was performed in a server with the following hardware: 4 × Intel(R) Xeon(R) CPU E5-4620 0 @ 2.20GHz processors, 8 cores per processor with HyperThreading, 10 TB HDD, 512 GB RAM. The server software configuration comprises Ubuntu 18.04 and R 3.4.4.

The source code of our proposal can be found in the online repository~\footnote{Complexity Measures and Features for Multivariate Times Series classification. \url{https://github.com/fjbaldan/CMFMTS/}}.

\subsection{Results}
\label{Results}

\begin{table}[]
	\centering
	\caption{Accuracy results on the UEA repository datasets: accuracy (\%), average accuracy, median, average rank, and Win/Loss/Tie Ratio. The best results are stressed in bold.}
	\label{totalTypeAcc}
	\begin{sideways}
	\resizebox{1.57\textwidth}{!}{
	\begin{tabular}{|l|l|l|l|l|l|l|l|l|l|l|l|l|l|l|l|}
		\toprule
		Datasets                  &  \makecell{CMFM+\\C5.0B} & \makecell{CMFM+\\RF}                    & \makecell{CMFM+\\SVM} & \makecell{CMFM+\\1NN-ED} & LCEM                          & XGBM                          & RFM                            & \makecell{MLSTM\\-\\FCN}                     & \makecell{WEASEL\\+\\MUSE}                   & \makecell{ED-\\1NN} & \makecell{DTW-\\1NN-I}                     & \makecell{DTW-\\1NN-D}                     & \makecell{ED-\\1NN\\(norm)} & \makecell{DTW-\\1NN-I\\(norm)}               & \makecell{DTW-\\1NN-D\\(norm)}                \\
		\midrule
		ArticularyWordRecognition & 91         & 99                              & 97.7        & 98.3           & \textbf{99.3} & 99                              & 99                              & 98.6                            & \textbf{99.3} & 97     & 98                              & 98.7                            & 97            & 98                              & 98.7                            \\
		AtrialFibrilation         & 20         & 20                              & 26.7        & 13.3           & \textbf{46.7} & 40                              & 33.3                            & 20                              & 26.7                            & 26.7   & 26.7                            & 20                              & 26.7          & 26.7                            & 22                              \\
		BasicMotions              & 85         & 97.5                            & 92.5        & 95             & \textbf{100} & \textbf{100} & \textbf{100} & \textbf{100} & \textbf{100} & 67.5   & \textbf{100} & 97.5                            & 67.6          & \textbf{100} & 97.5                            \\
		CharacterTrajectories     & 95.3       & 97.1                            & 95.9        & 90.7           & 97.9                            & 98.3                            & 98.5                            & \textbf{99.3} & 99                              & 96.4   & 96.9                            & 99                              & 96.4          & 96.9                            & 98.9                            \\
		Cricket                   & 86.1       & 97.2                            & 95.8        & 97.2           & 98.6                            & 97.2                            & 98.6                            & 98.6                            & 98.6                            & 94.4   & 98.6                            & \textbf{100} & 94.4          & 98.6                            & \textbf{100} \\
		DuckDuckGeese             & 42         & 52                              & 44          & 40             & 37.5                            & 40                              & 40                              & \textbf{67.5} & 57.5                            & 27.5   & 55                              & 60                              & 27.5          & 55                              & 60                              \\
		EigenWorms                & 81.7       & 88.5                            & 84          & 79.4           & 52.7                            & 55                              & \textbf{100} & 80.9                            & 89                              & 55     & 60.3                            & 61.8                            & 54.9          & 61.8                            & NA                              \\
		Epilepsy                  & 88.4       & \textbf{100} & 97.8        & 95.7           & 98.6                            & 97.8                            & 98.6                            & 96.4                            & 99.3                            & 66.7   & 97.8                            & 96.4                            & 66.6          & 97.8                            & 96.4                            \\
		Ering                     & 80.7       & \textbf{93} & 92.6        & 90.4           & 20                              & 13.3                            & 13.3                            & 13.3                            & 13.3                            & 13.3   & 13.3                            & 13.3                            & 13.3          & 13.3                            & 13.3                            \\
		EthanolConcentration      & 35         & 33.5                            & 32.7        & 30.4           & 37.2                            & 42.2                            & \textbf{43.3} & 27.4                            & 31.6                            & 29.3   & 30.4                            & 32.3                            & 29.3          & 30.4                            & 32.3                            \\
		FaceDetection             & 54         & 54.8                            & 57.9        & 50.5           & 61.4                            & \textbf{62.9} & 61.4                            & 55.5                            & 54.5                            & 51.9   & 51.3                            & 52.9                            & 51.9          & 52.9                            & NA                              \\
		FingerMovements           & 44         & 52                              & 46          & 53             & 59                              & 53                              & 56                              & \textbf{61} & 54                              & 55     & 52                              & 53                              & 55            & 52                              & 53                              \\
		HandMovementDirection     & 33.8       & 28.4                            & 32.4        & 18.9           & \textbf{64.9} & 54.1                            & 50                              & 37.8                            & 37.8                            & 27.9   & 30.6                            & 23.1                            & 27.8          & 30.6                            & 23.1                            \\
		Handwriting               & 16.5       & 28.2                            & 18.4        & 24.9           & 28.7                            & 26.7                            & 26.7                            & 54.7                            & 53.1                            & 37.1   & 50.9                            & \textbf{60.7} & 20            & 31.6                            & 28.6                            \\
		Heartbeat                 & 74.1       & 76.6                            & 73.2        & 62             & 76.1                            & 69.3                            & \textbf{80} & 71.4                            & 72.7                            & 62     & 65.9                            & 71.7                            & 61.9          & 65.8                            & 71.7                            \\
		InsectWingbeat            & NA         & \textbf{64.0} & 10          & 25.8           & 22.8                            & 23.7                            & 22.4                            & 10.5                            & 12.8                            & 11.5   & 12.8                            & NA                              & NA            & NA                              & NA                              \\
		JapaneseVowels            & 82.4       & 87.6                            & 76.5        & 72.2           & 97.8                            & 96.8                            & 97                              & \textbf{99.2} & 97.8                            & 92.4   & 95.9                            & 94.9                            & 92.4          & 95.9                            & 94.9                            \\
		Libras                    & 83.9       & 86.7                            & 83.3        & 82.8           & 77.2                            & 76.7                            & 78.3                            & \textbf{92.2} & 89.4                            & 83.3   & 89.4                            & 87.2                            & 83.3          & 89.4                            & 87                              \\
		LSST                      & 63.1       & \textbf{65.2} & 64.8        & 50             & \textbf{65.2} & 63.3                            & 61.2                            & 64.6                            & 62.8                            & 45.6   & 57.5                            & 55.1                            & 45.6          & 57.5                            & 55.1                            \\
		MotorImagery              & 49         & 51                              & 50          & 44             & \textbf{60.0} & 46                              & 55                              & 53                              & 50                              & 51     & 39                              & 50                              & 51            & 50                              & NA                              \\
		NATOPS                    & 81.7       & 81.7                            & 75          & 73.9           & 91.6                            & 90                              & 91.1                            & \textbf{96.1} & 88.3                            & 85     & 85                              & 88.3                            & 85            & 85                              & 88.3                            \\
		PenDigits                 & 93.3       & 95.1                            & 95.9        & 94.4           & 97.7                            & 95.1                            & 95.1                            & \textbf{98.7} & 96.9                            & 97.3   & 93.9                            & 97.7                            & 97.3          & 93.9                            & 97.7                            \\
		PEMSF                     & 96.5       & \textbf{100} & 69.4        & 77.5           & 94.2                            & 98.3                            & 98.3                            & 65.3                            & 70.5                            & 73.4   & 71.1                            & 70.5                            & 73.4          & 71.1                            & NA                              \\
		PhonemeSpectra                   & 22.4       & 28.7                            & 25          & 15.8           & \textbf{28.8} & 18.7                            & 22.2                            & 27.5                            & 19                              & 10.4   & 15.1                            & 15.1                            & 10.4          & 15.1                            & 15.1                            \\
		RacketSports              & 72.4       & 80.9                            & 80.9        & 71.1           & \textbf{94.1} & 92.8                            & 92.1                            & 88.2                            & 91.4                            & 86.4   & 84.2                            & 80.3                            & 86.8          & 84.2                            & 80.3                            \\
		SelfRegulationSCP1        & 81.2       & 81.2                            & 79.2        & 70.3           & 83.9                            & 82.9                            & 82.6                            & \textbf{86.7} & 74.4                            & 77.1   & 76.5                            & 77.5                            & 77.1          & 76.5                            & 77.5                            \\
		SelfRegulationSCP2        & 53.9       & 41.7                            & 46.1        & 50             & \textbf{55} & 48.3                            & 47.8                            & 52.2                            & 52.2                            & 48.3   & 53.3                            & 53.9                            & 48.3          & 53.3                            & 53.9                            \\
		SpokenArabicDigits        & 93.9       & 96.9                            & 97.5        & 92.6           & 97.3                            & 97                              & 96.8                            & \textbf{99.4} & 98.2                            & 96.7   & 96                              & 96.3                            & 96.7          & 95.9                            & 96.3                            \\
		StandWalkJump             & 26.7       & 33.3                            & 20          & 13.3           & 40                              & 33.3                            & \textbf{46.7} & \textbf{46.7} & 33.3                            & 20     & 33.3                            & 20                              & 20            & 33.3                            & 20                              \\
		UWaveGestureLibrary       & 64.1       & 77.2                            & 73.8        & 75.3           & 89.7                            & 89.4                            & 90                              & 85.7                            & \textbf{90.3} & 88.1   & 86.9                            & \textbf{90.3} & 88.1          & 86.8                            & \textbf{90.3} \\
		\midrule
		Mean              & 63.1       & \textbf{69.6}                            & 64.5        & 61.6           & 69.1                            & 66.7                            & 69.2                            & 68.3                            & 67.1                            & 59.1   & 63.9                            & 63.9                            & 58.2          & 63.3                            & 55.1                            \\
		Median             & 73.3       & 79.1                            & 73.5        & 70.7           & 70.7                            & 66.3                            & \textbf{79.2}                            & 69.5                            & 71.6                            & 58.5   & 63.1                            & 66.2                            & 58.5          & 63.8                            & 57.6                            \\
		Average Rank               & 9.95       & 6.88                             & 9.03         & 11.53           & \textbf{4.1}                             & 6.57                             & 5.15                             & 5.3                             & 5.62                             & 10.37   & 8.83                             & 7.9                             & 10.9          & 8.9                             & 8.97                             \\
		Win/Loss/Tie Ratio        & 0/30/0     & 5/25/1                          & 0/30/0      & 0/30/0         & 9/21/3                          & 2/28/1                          & 5/25/2                          & \textbf{11/19/2}                         & 3/27/3                          & 0/30/0 & 1/29/1                          & 3/27/2                          & 0/30/0        & 1/29/1                          & 2/28/2     \\                    
		\bottomrule	
	\end{tabular}
}
\end{sideways}
\end{table}

We start by analyzing the accuracy and the average rank results on the 30 processed datasets. Table~\ref{totalTypeAcc} shows the accuracy results obtained by our proposal against the main state-of-the-art algorithms. The NA values refer to cases in which for any reason (memory overflow, limitation of libraries, etc.) a model has not been obtained correctly, and it has been impossible to perform the desired classification. As we see in Table~\ref{totalTypeAcc}, our proposal CMFMTS+RF, called CMFM+RF for simplification, obtains the best results among the four models we have proposed: CMFMTS+C5.0B, CMFMTS+RF, CMFMTS+SVM, and CMFMTS+1NN-ED. The CMFMTS+C5.0B model is especially interesting for cases where a simple classifier is required, easy to interpret, and that offers results close to the optimum ones, as it happens in the datasets: CharactertTrajectories, LSST, and PEMSF. We can find cases in which CMFMTS+RF does not offer the best results among these four models, and it may be interesting to try other combinations as it happens in the datasets: AtrialFibrilation, EthanolConcentration, FaceDetection, FingerMovements, etc.

\begin{figure}[!ht]
	\includegraphics[width=\textwidth]{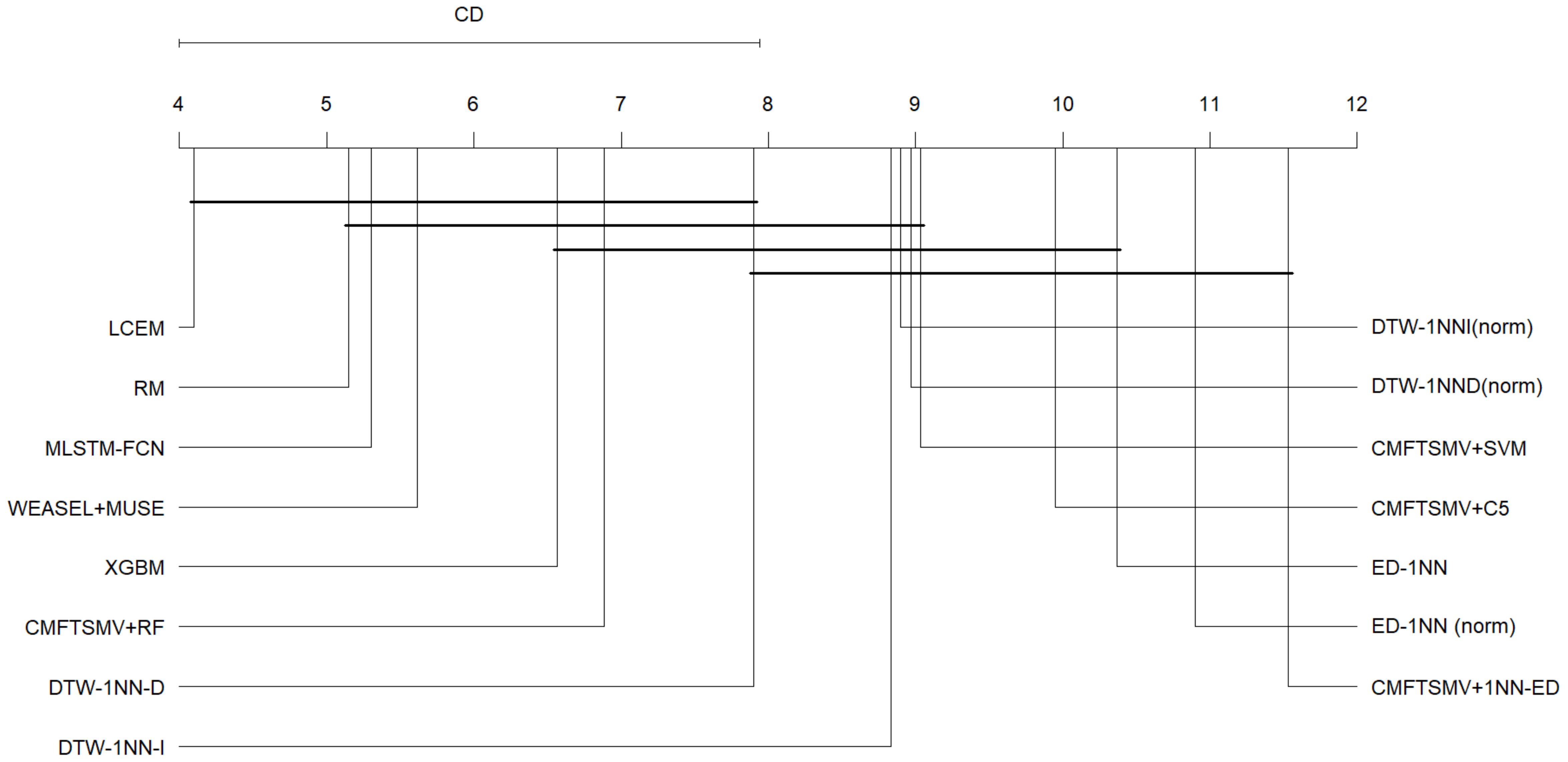}
	\caption{Critical Difference diagram, $\alpha = 0.05$.}
	\label{fig:CDDiagram}
\end{figure}

If we compare our proposal with the rest of the state-of-the-art algorithms we see how CMFMTS+RF obtains an average rank of 6.88, close to the one obtained by LCEM (4.1), RFM (5.15), MLSTM-FCN (5.3), and WEASEL+MUSE (5.62). We have included two decimals for the average rank so that the differences shown in Figure~\ref{fig:CDDiagram} can be better appreciated. If we observe the Win/Loss/Tie ratio, we can see that MLSTM-FCN obtains the best results in 11 datasets, followed by LCEM which wins in 9 datasets, and CMFMTS+RF and RFM, which obtains the best results in 5 datasets. These behaviors are reflected in the CD diagram shown in Figure~\ref{fig:CDDiagram}. This diagram shows that there is no statistically significant difference, for an $\alpha$ of 0.05, between the previously mentioned models, in addition to the DTW-1NN-D model. This indicates that our CMFMTS+RF proposal offers results that are statically indistinguishable from those obtained by the main state-of-the-art algorithms. Analyzing the median and average values of accuracy our proposal obtains competitive results. The NA values have been transformed to 0 for the calculation of these last two measurements.

Analyzing the results obtained for some specific cases, we can appreciate significant differences between the different proposals. For example, in the DuckDuckGeese dataset, the MLSTM-FCN algorithm obtains 7.5 points of difference with the next best result. LCEM and similar proposals obtain results with significant differences concerning the rest of the methods, as can be seen in the HandMovementDirection dataset. In the Ering dataset, we see a big difference between our CMFMTS+Any proposals and the rest of the algorithms. These cases confirm the idea that in the field of CMTS the results are strongly linked to the data itself and the approach used. It is especially complicated to find an approach able to face all kinds of problems with optimal results, or close to them.

The difference in results between the traditional TSC approach and our proposal is largely due to the nature of the underlying problem. In general, it can be seen that the features extracted from the time series summarize behavior at the general level of the time series. Although the classifiers used are capable of finding relationships of interest between the features of the different variables, there are cases in which the class differentiator may fall entirely on specific patterns not reflected in those features. In these cases traditional MTSC approaches such as 1NN-DTW can directly find such patterns.

\subsection{Analysis and interpretability of results}
\label{Interpretability}
The interpretability of the final results is strongly related to the model used, for example, the C5.0B model offers us a simple decision tree based on the features used, although as we saw in Table~\ref{totalTypeAcc} its accuracy results are not the best. On the other hand, a RF offers competitive results in exchange for sacrificing part of their interpretability, although RF is able to offer an assessment of the importance of each feature in the final model that can be very useful. In contrast, models such as 1NN-ED lack interpretability, since they work on how much one instance resembles another, and SVMs are really complex to interpret since weights can be affected by external components unrelated to the underlying importance of each variable. Since tree-based models offer different interpretability tools, we will focus on them in this section.

\renewcommand{\thesubfigure}{\alph{subfigure}}
\begin{figure}[!t]
	\centering
	\begin{subfigure}[b]{\textwidth}
		\includegraphics[width=\textwidth]{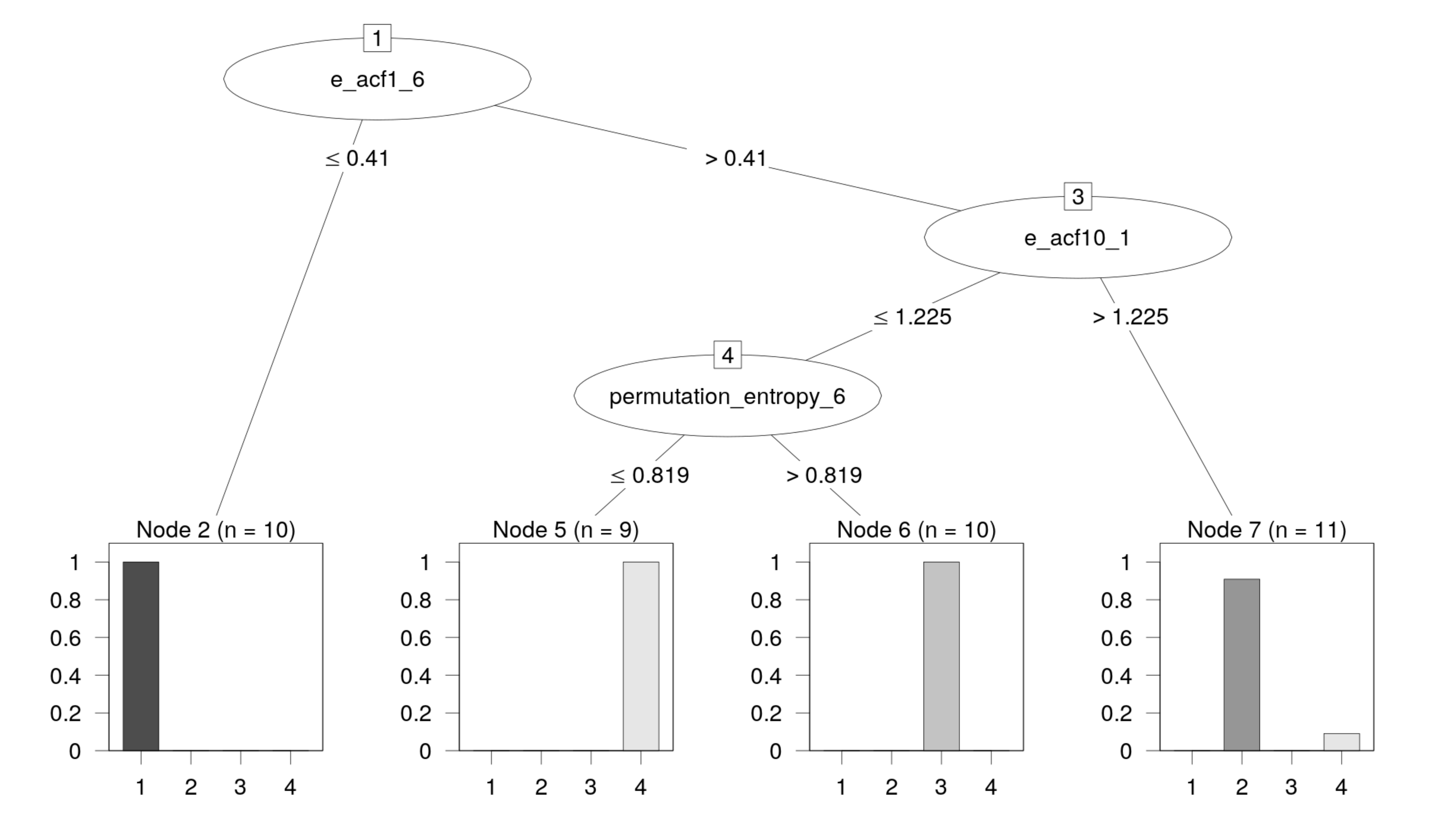}
		\caption{BasicMotions example of a single C5.0B tree with time series features.}
		\label{fig:exampleC5:a}
	\end{subfigure}
	\begin{subfigure}[b]{\textwidth}
		\includegraphics[width=\textwidth]{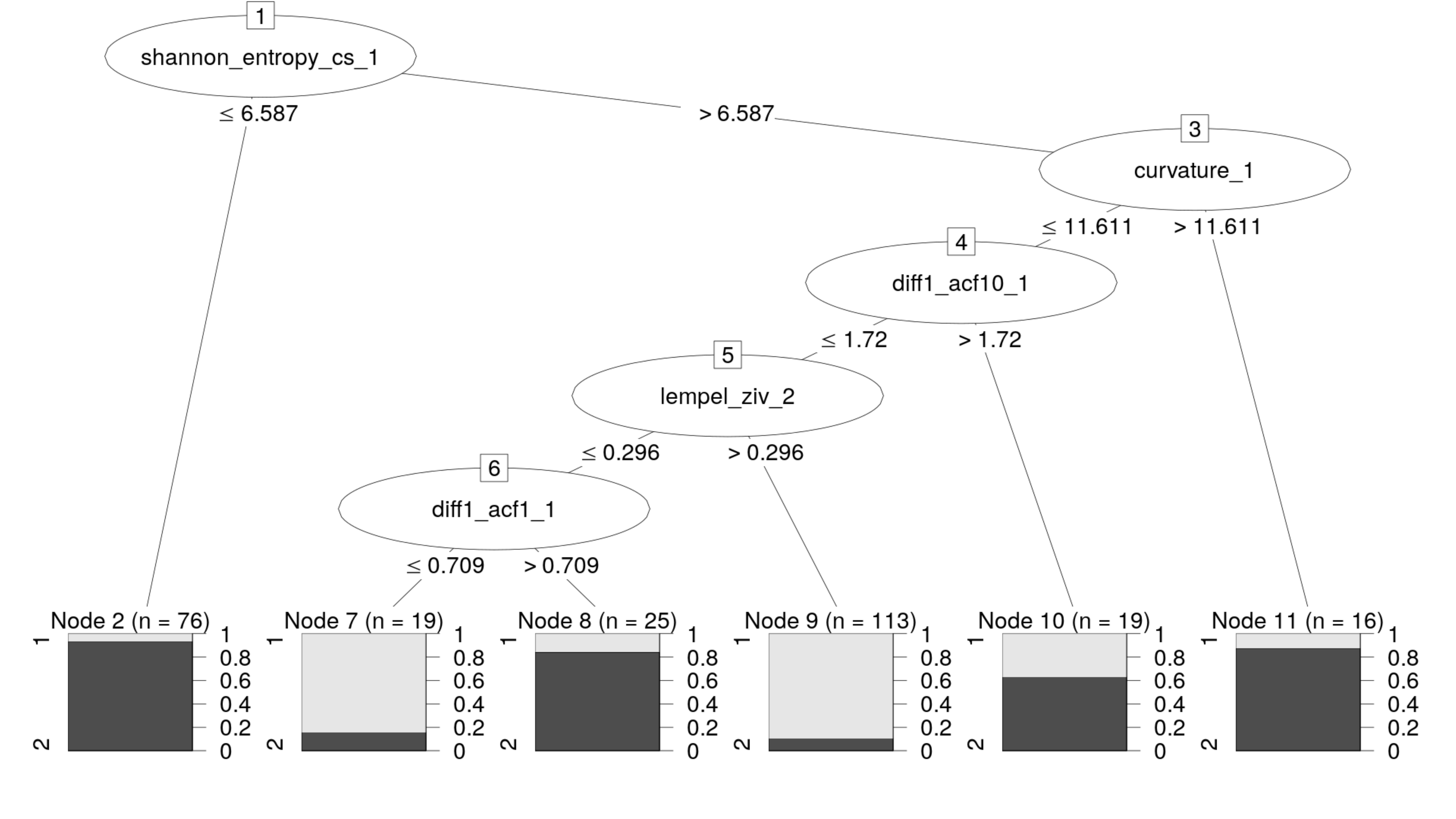}
		\caption{SelfRegulationSCP1 example of a single C5.0B tree with time series features.}
		\label{fig:exampleC5:b}
	\end{subfigure}
	
	\caption{Interpretability CMF example.}
	\label{fig:exampleC5}
\end{figure}

%
%
%
In Figure~\ref{fig:exampleC5}, we show two examples of a single C5.0B trees for the BasicMotions and SelfRegulationSCP1 datasets. BasicMotions is a dataset with 4 classes and MTS with 6 variables. As we see in Figure~\ref{fig:exampleC5:a}, our approach allows us to solve this problem with a simple tree composed of 3 nodes. Two of these nodes refer to characteristics of variable 6 and the remaining one refers to variable 1. Although the results obtained for this dataset are not competitive, it is remarkable how you can obtain acceptable results with such a simple decision tree. The dataset SelfRegulationSCP1 is a binary MTS problem with 6 variables. In Figure~\ref{fig:exampleC5:b}, we see how 5 nodes have been necessary, 4 refer to variable 1, and the remaining to variable 2. In this case, the results obtained, although they are not the best, are close to the most competitive models. In both cases, although we have 6 variables per problem, we can see that the created C5.0B models focus on using information from only 2 variables. This fact offers us a slight confirmation about a typical phenomenon of MTS, and that is that not all variables of an MTS have the same importance.

In the case of the RF we have the Mean Decrease Gini Importance, which indicates the importance of each feature in a specific dataset. This information can be used to improve the learning process and better understand the problem. For this reason, we perform three different analyses that allow us to extract the desired information:

\begin{enumerate}
	\item Analysis of the importance of each feature in each dataset (Section~\ref{fi}). This allows us to know which features have a greater contribution to the final result. Based on the features with the greatest contribution, we can determine which behaviors, represented in those features, are the ones that define each type of time series.
	\item Analysis of the accumulated importance of each feature over a large set of datasets (Section~\ref{afid}). This analysis allows us to identify which features are representative of most problems and which ones are uninteresting.
	\item Analysis focused on the cumulative importance of features for each variable composing the MTS (Section~\ref{vi}). In this way, we could identify which variables contain a greater amount of information about a given problem. These variables would be the most interesting ones to solve the problem.
\end{enumerate}

\subsubsection{Feature importance by dataset}
\label{fi}
To analyze in a simple way the importance of the features used, we have chosen a graphic approach.

\begin{figure}[!ht]
	\includegraphics[width=\textwidth]{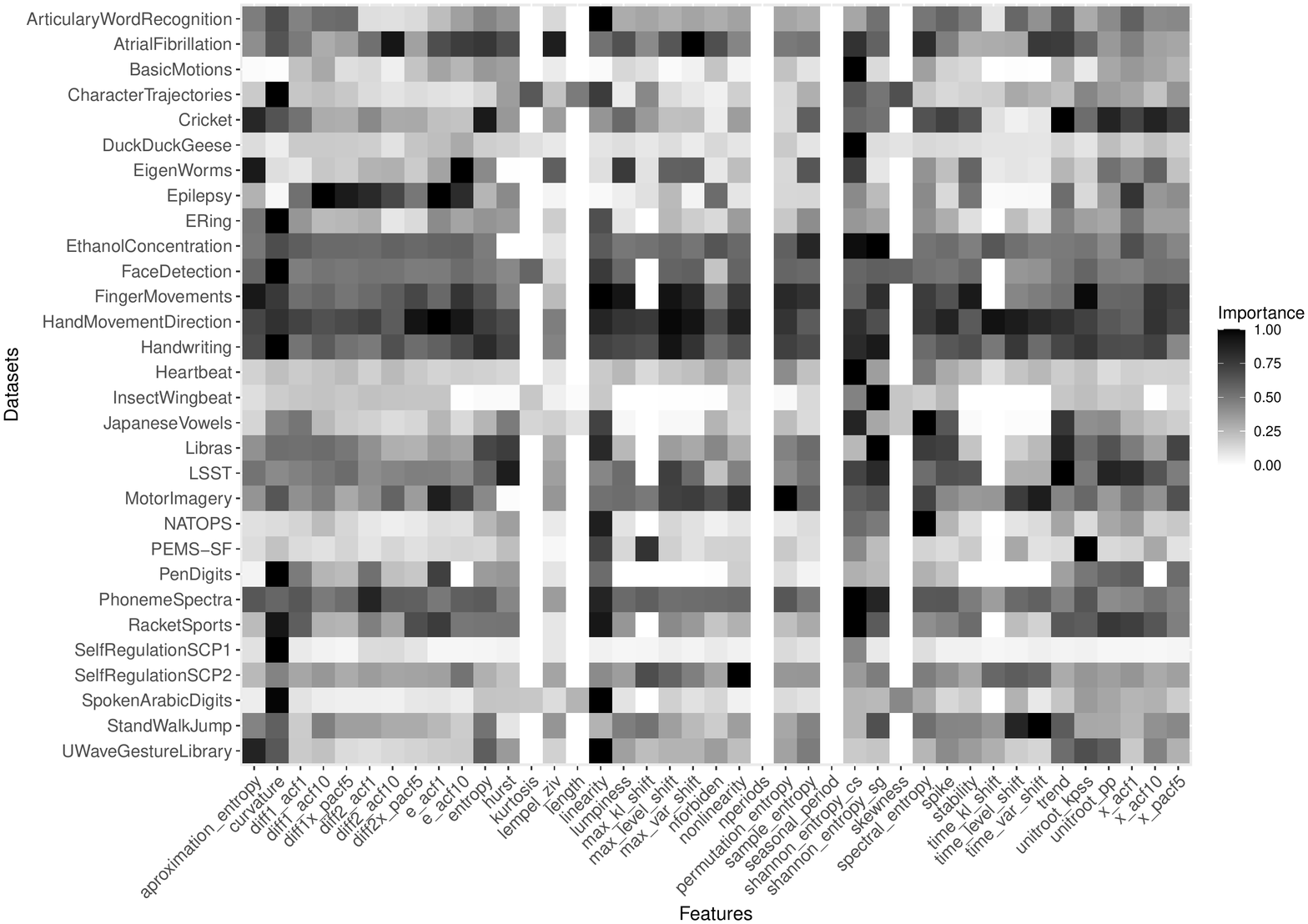}
	\caption{Features importance heatmap.}
	\label{fig:heatmap}
\end{figure}

Figure~\ref{fig:heatmap} shows a heatmap of the importance of each feature in the RF classifier for each dataset. This importance has been obtained by accumulating the value of Mean Decrease Gini Importance of each feature in the different variables of each MTS. For example, for the approximation entropy feature in a 7-variables MTS, we get 7 different values of importance (1 for each variable to which its corresponding approximation entropy feature is calculated). We add these 7 values and divide them by the number of variables of our MTS. In this way, we also penalize the importance of any features that could not be calculated in any variable. Finally, we normalize these last values between 0 and 1 for each dataset.

In Figure~\ref{fig:heatmap}, we see significant differences among the different datasets. First, we see datasets with solutions dominated by 1 to 4 features with high importance (BasicMotions, DuckDuckGeese, Heartbeat, InsectWingBeat, NATOPS, PEMS-SF, SelfRegulationSCP1, and SpokenArabicDigits). In these cases, we have two options: the set of features used is sufficiently expressive to address the problem satisfactorily with competitive results (RF: Heartbeat, InsectWingBeat, and PEMS-SF) or the selected features are not sufficient and other approaches achieve significantly better results (MLSTM-FCN: BasicMotions, DuckDuckGeese, NATOPS, SelfRegulationSCP1, and SpokenArabicDigits).

Secondly, we can identify cases where all the features are necessary (AtrialFibrillation, EthanolConcentration, FingerMovements, HandMovementDirection, Handwriting, LSST, MotorImagery, and PhonemeSpectra). We think there are two possible explanations for this situation. Firstly, in some cases not even the whole set of 41 features provide enough information to produce an accurate classifier. For this reason, the classifier assigns similar importance to a large number of features and is not able to obtain the best results (AtrialFibrillation, EthanolConcentration, FingerMovements, HandMovementDirection, Handwriting, and MotorImagery). Secondly, complex cases in which it is not possible to find a reduced subset of features capable of explaining the problem. In these cases, more complex solutions are obtained, with a high number of features, capable of offering the best results (LSST and PhonemeSpectra, for this last dataset we obtain results very close to the best ones).

\begin{figure}[!ht]
	\includegraphics[width=\textwidth]{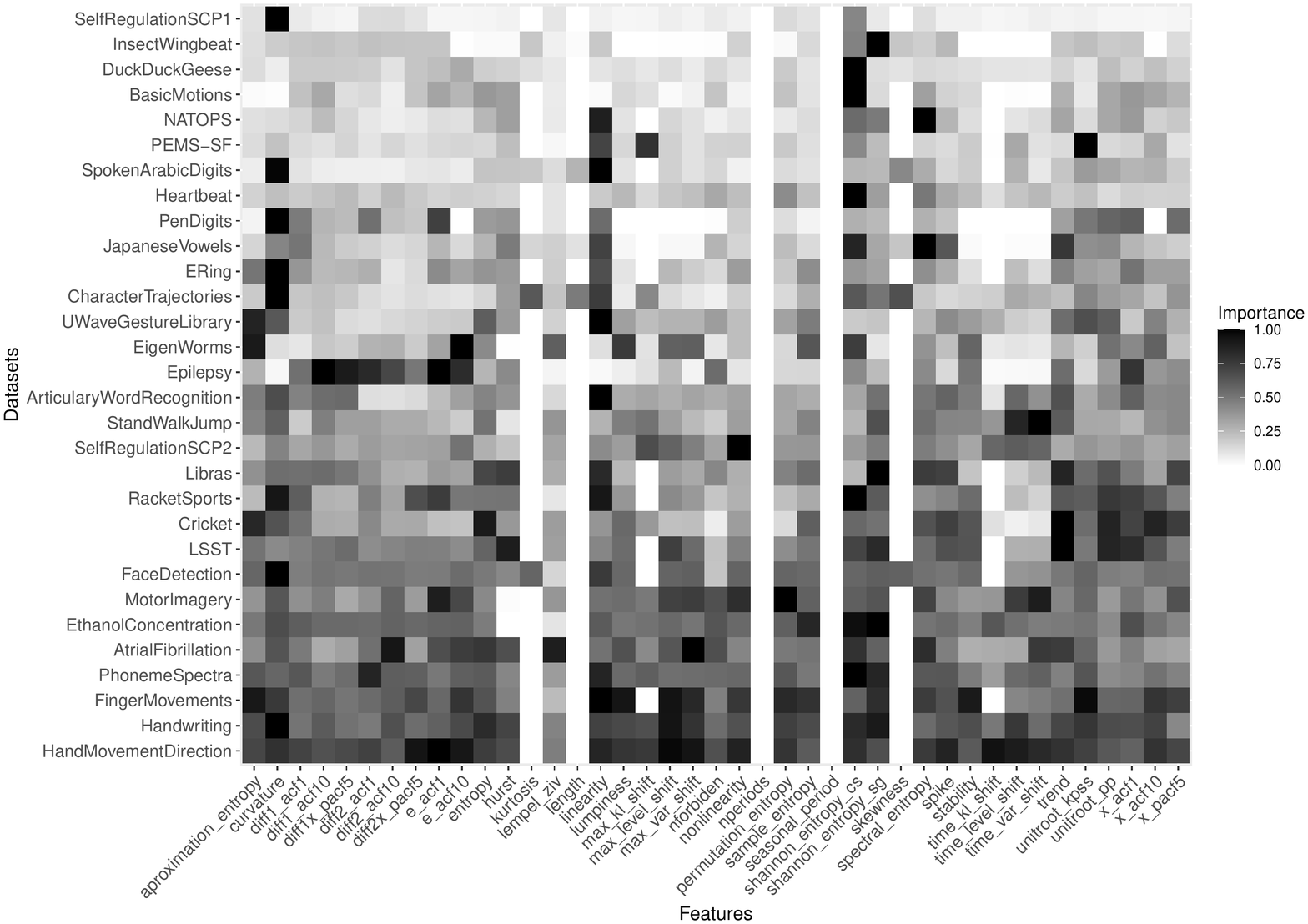}
	\caption{Ordered features importance heatmap.}
	\label{fig:heatmapOrder}
\end{figure}

These behaviors and the differences mentioned above can be grasped clearly from Figure~\ref{fig:heatmapOrder}, where the datasets have been ordered by the accumulated importance of the 41 features. If we look from bottom to top, the cases from highest to lowest accumulated importance, we see how from HandMovementDirection to ArticularyWordRecognition, from 15 datasets in only 1 case, LSST, our proposal featured-based achieves the best result. On the other hand, if we look from SelfRegulationSCP1 to Epilepsy, we see that our proposal obtains the best results in 4 out of 15 cases. So we can infer that in cases where great importance is given to a high number of features, this approach does not lead to the best results.

\subsubsection{Accumulated feature importance over a large set of datasets}
\label{afid}

\begin{figure}[!ht]
	\includegraphics[width=\textwidth]{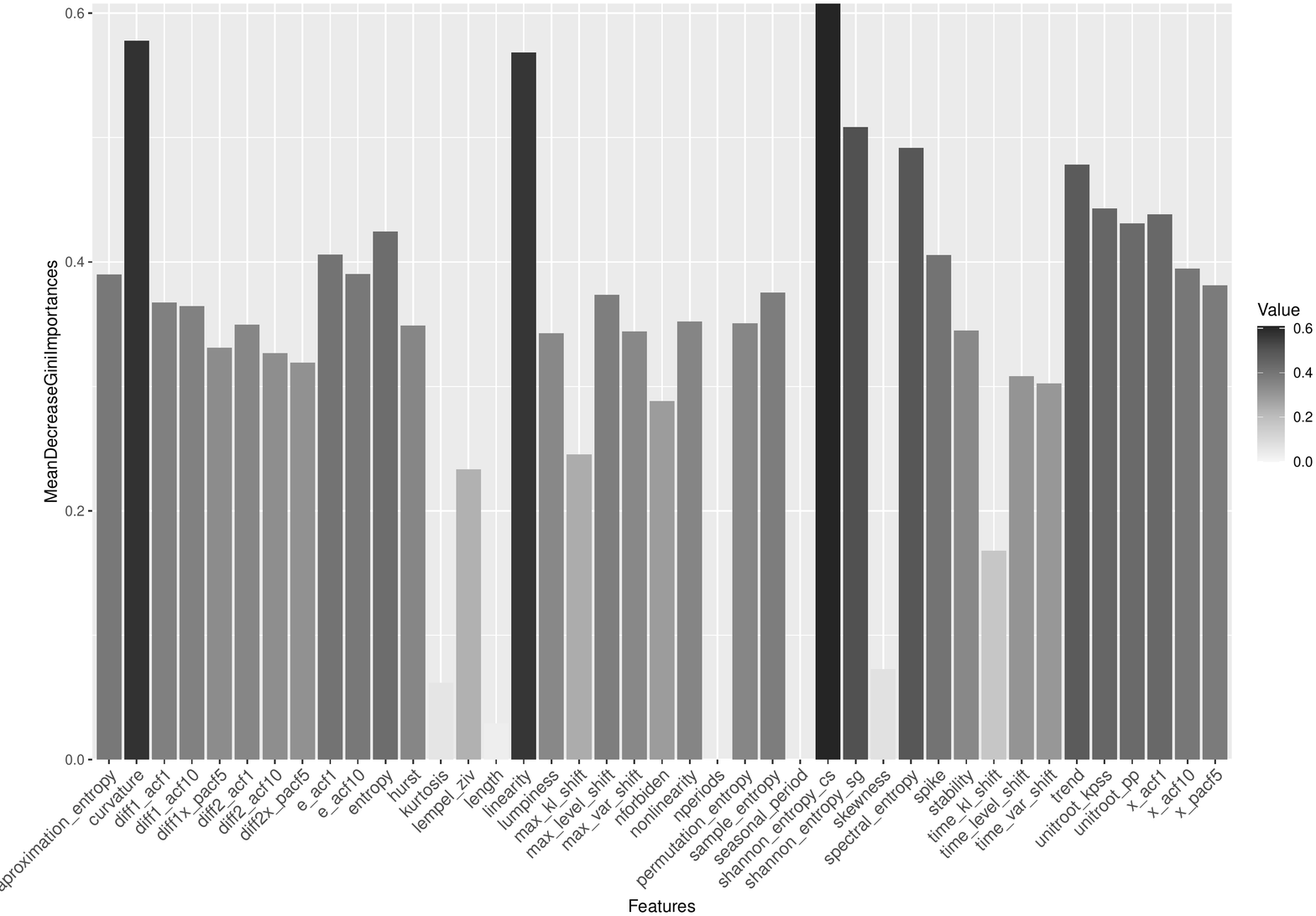}
	\caption{Average importance of each feature in the UEA repository.}
	\label{fig:impByFeature}
\end{figure}

Another particularly interesting analysis to be carried out is related to the importance at the feature level. In Figure~\ref{fig:impByFeature}, we show the average importance of each feature throughout all the datasets. These values have been obtained from the results shown in Figures~\ref{fig:heatmap} and~\ref{fig:heatmapOrder}. The average value of the importance of each feature has been calculated over the 30 datasets processed. We can see that there is a group of distinguished features, namely, \textit{curvature}, \textit{linearity}, and \textit{shannon\_entropy\_cs}. This group has values of importance far superior to the rest. There is also a second group with high values of importance but far from the highest values: \textit{shannon\_entropy\_sg}, \textit{spectral\_entropy}, \textit{trend}, \textit{unitroot\_kpss}, \textit{unitroot\_pp}, \textit{x\_acf1}, \textit{entropy}, \textit{e\_acf1}, and \textit{spike}. All of them are assigned average importance values much higher than the rest of features, greater than 0.4. Even further, it is interesting to realize the features related to the complexity of a time series get the highest importance values. Higher values achieved by features such as \textit{trend}, \textit{x\_acf1}, and \textit{e\_acf1} confirm that the components of the time series are very descriptive and useful when extracting information from them. Other features such as \textit{curvature}, \textit{linearity}, and \textit{spike}, shown as characteristic behaviors of the time series, are especially useful in describing them.

On the other hand, there are also features with particularly low values of importance. In this case, the time series have been treated as traditional data vectors, for that reason features such as \textit{nperiods} and \textit{seasonal\_periods} have an importance values of 0. In the case that the best results are sought and a detailed analysis of the time series is carried out, in which data on seasonality are available for each time series, these measures can be very useful. In the UEA repository, the vast majority of datasets are composed of MTS of equal length, for this reason, the \textit{length} feature is not of high importance. Furthermore, features such as \textit{kurtosis} and \textit{skewness} have obtained low average importance values, although they are particularly explanatory. If we look at Figures~\ref{fig:heatmap} and~\ref{fig:heatmapOrder}, we find some datasets like FaceDetection, CharacterTrajectories, SpokenArabicDigits, and InsectWingbeat in which these features have obtained high importance values. In this case, even if we identify features that are generally not interesting, they may be relevant for specific problems. These cases reinforce the idea that the selection of a representative set of features must be supported by theoretical knowledge about the structure of time series and by different analyses of results performed on large sets of datasets.

\subsubsection{Variable importance}
\label{vi}
Finally, we analyze an important point in the field of MTSC, the existence of components or variables that contain a greater part of the information on the problem. To do this, we have calculated the sum of the 41 features for each variable of the problem in question. Then we have rescaled these values by dividing by the maximum value of each case. In this way, the maximum importance value of any variable, in any case, will be 1. The range of possible importance values is [0,1]. In Table~\ref{statistics}, we show the statistics of interest on the values obtained.

\begin{table}[!t]
	\centering
	\caption{Statistics of the accumulated importance value by variable.}
	\label{statistics}
	\resizebox{1\textwidth}{!}{
	\begin{tabular}{|l|l|l|l|l|l|l|l|l|}
		\toprule
		 Datasets                    &  Sum   &  Max  &  Min  &  Mean  &  Median  &  Var  &  SD   &  Variables  \\
		 \midrule
		 ArticularyWordRecognition & 4.876   & 1   & 0.196 & 0.542 & 0.518  & 0.073 & 0.271 & 9    \\
		 AtrialFibrillation        & 1.839   & 1   & 0.839 & 0.92  & 0.92   & 0.013 & 0.114 & 2    \\
		 BasicMotions              & 3.362   & 1   & 0.215 & 0.56  & 0.471  & 0.139 & 0.372 & 6    \\
		 CharacterTrajectories     & 2.212   & 1   & 0.406 & 0.737 & 0.806  & 0.092 & 0.303 & 3    \\
		 Cricket                   & 5.038   & 1   & 0.661 & 0.84  & 0.845  & 0.017 & 0.131 & 6    \\
		 DuckDuckGeese             & 141.067 & 1   & 0     & 0.105 & 0.076  & 0.013 & 0.112 & 1345 \\
		 EigenWorms                & 3.922   & 1   & 0.467 & 0.654 & 0.618  & 0.032 & 0.18  & 6    \\
		 Epilepsy                  & 2.243   & 1   & 0.345 & 0.748 & 0.898  & 0.124 & 0.352 & 3    \\
		 EthanolConcentration      & 2.954   & 1   & 0.96  & 0.985 & 0.994  & 0     & 0.021 & 3    \\
		 ERing                     & 3.314   & 1   & 0.714 & 0.829 & 0.8    & 0.016 & 0.128 & 4    \\
		 FaceDetection             & 95.112  & 1   & 0.548 & 0.66  & 0.659  & 0.004 & 0.062 & 144  \\
		 FingerMovements           & 20.031  & 1   & 0.597 & 0.715 & 0.689  & 0.009 & 0.093 & 28   \\
		 HandMovementDirection     & 9.069   & 1   & 0.808 & 0.907 & 0.908  & 0.004 & 0.067 & 10   \\
		 Handwriting               & 2.692   & 1   & 0.806 & 0.897 & 0.885  & 0.009 & 0.097 & 3    \\
		 Heartbeat                 & 30.381  & 1   & 0.324 & 0.498 & 0.463  & 0.019 & 0.137 & 61   \\
		 InsectWingbeat            & 71.798  & 1   & 0.251 & 0.359 & 0.284  & 0.023 & 0.151 & 200  \\
		 JapaneseVowels            & 8.115   & 1   & 0.458 & 0.676 & 0.678  & 0.025 & 0.158 & 12   \\
		 Libras                    & 1.877   & 1   & 0.877 & 0.939 & 0.939  & 0.008 & 0.087 & 2    \\
		 LSST                      & 5.11    & 1   & 0.71  & 0.852 & 0.844  & 0.021 & 0.145 & 6    \\
		 MotorImagery              & 52.961  & 1   & 0.617 & 0.828 & 0.817  & 0.007 & 0.085 & 64   \\
		 NATOPS                    & 9.722   & 1   & 0.155 & 0.405 & 0.347  & 0.051 & 0.225 & 24   \\
		 PenDigits                 & 1.666   & 1   & 0.666 & 0.833 & 0.833  & 0.056 & 0.236 & 2    \\
		 PEMS-SF                   & 33.012  & 1   & 0     & 0.034 & 0.014  & 0.006 & 0.079 & 963  \\
		 PhonemeSpectra            & 10.802  & 1   & 0.969 & 0.982 & 0.982  & 0     & 0.009 & 11   \\
		 RacketSports              & 4.467   & 1   & 0.307 & 0.744 & 0.791  & 0.064 & 0.254 & 6    \\
		 SelfRegulationSCP1        & 4.138   & 1   & 0.582 & 0.69  & 0.627  & 0.025 & 0.158 & 6    \\
		 SelfRegulationSCP2        & 6.683   & 1   & 0.92  & 0.955 & 0.948  & 0.001 & 0.032 & 7    \\
		 SpokenArabicDigits        & 6.042   & 1   & 0.169 & 0.465 & 0.375  & 0.086 & 0.293 & 13   \\
		 StandWalkJump             & 3.319   & 1   & 0.71  & 0.83  & 0.805  & 0.018 & 0.132 & 4    \\
		 UWaveGestureLibrary       & 2.72    & 1   & 0.847 & 0.907 & 0.873  & 0.007 & 0.082 & 3	\\
		 \bottomrule
	\end{tabular}
	}

\end{table}

\begin{figure}[!ht]  
	\captionsetup[subfigure]{justification=centering}
	\centering
	\begin{subfigure}{0.49\textwidth}
		\includegraphics[width=\textwidth]{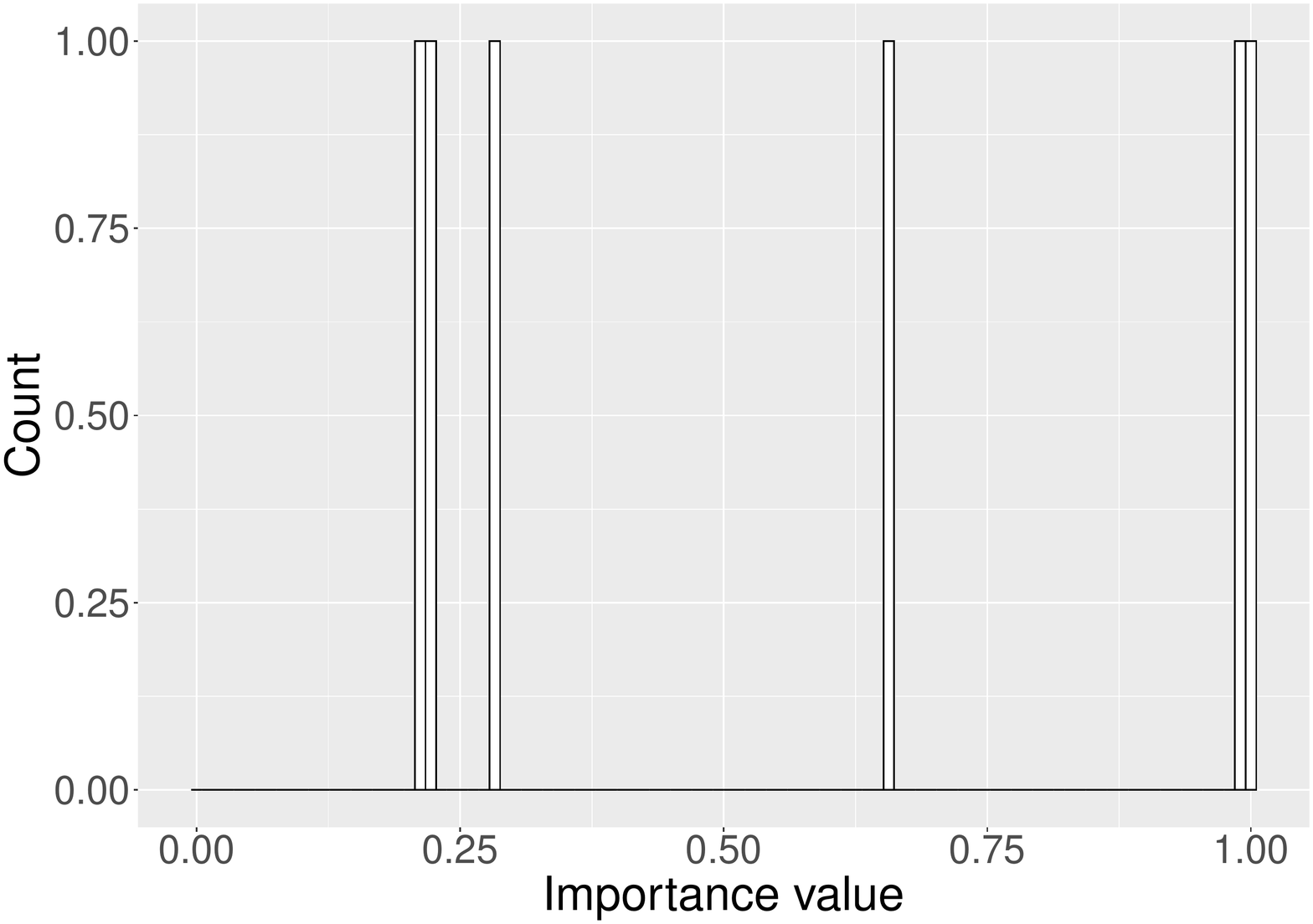}
		\caption{BasicMotions dataset.}
		\label{fig:hist:a}
	\end{subfigure}
	\hfill 
	\begin{subfigure}{0.49\textwidth}
		\includegraphics[width=\textwidth]{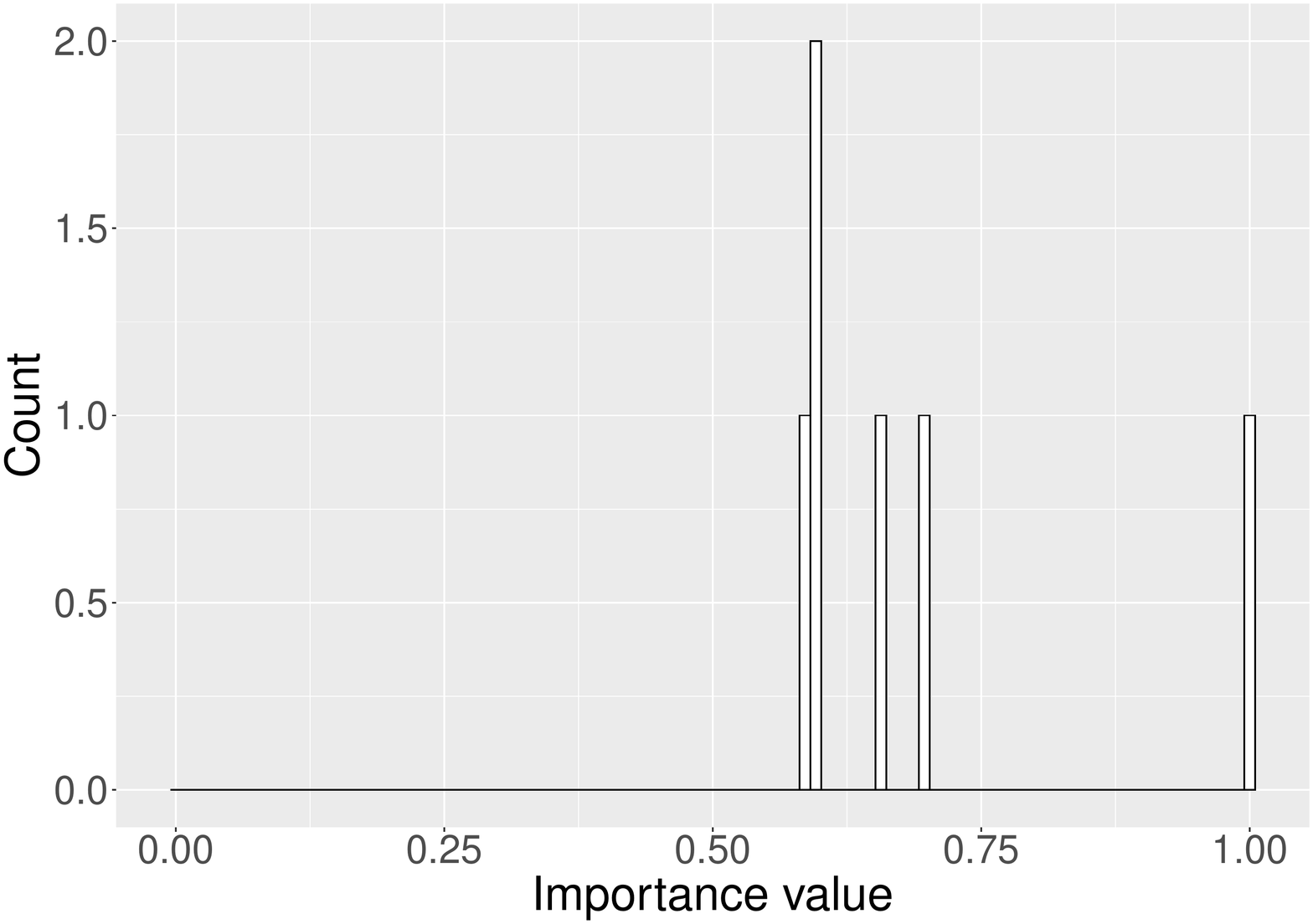}
		\caption{SelfRegulationSCP1 dataset.}
		\label{fig:hist:b}
	\end{subfigure}
	
	\bigskip  
	\begin{subfigure}{0.49\textwidth}
		\includegraphics[width=\textwidth]{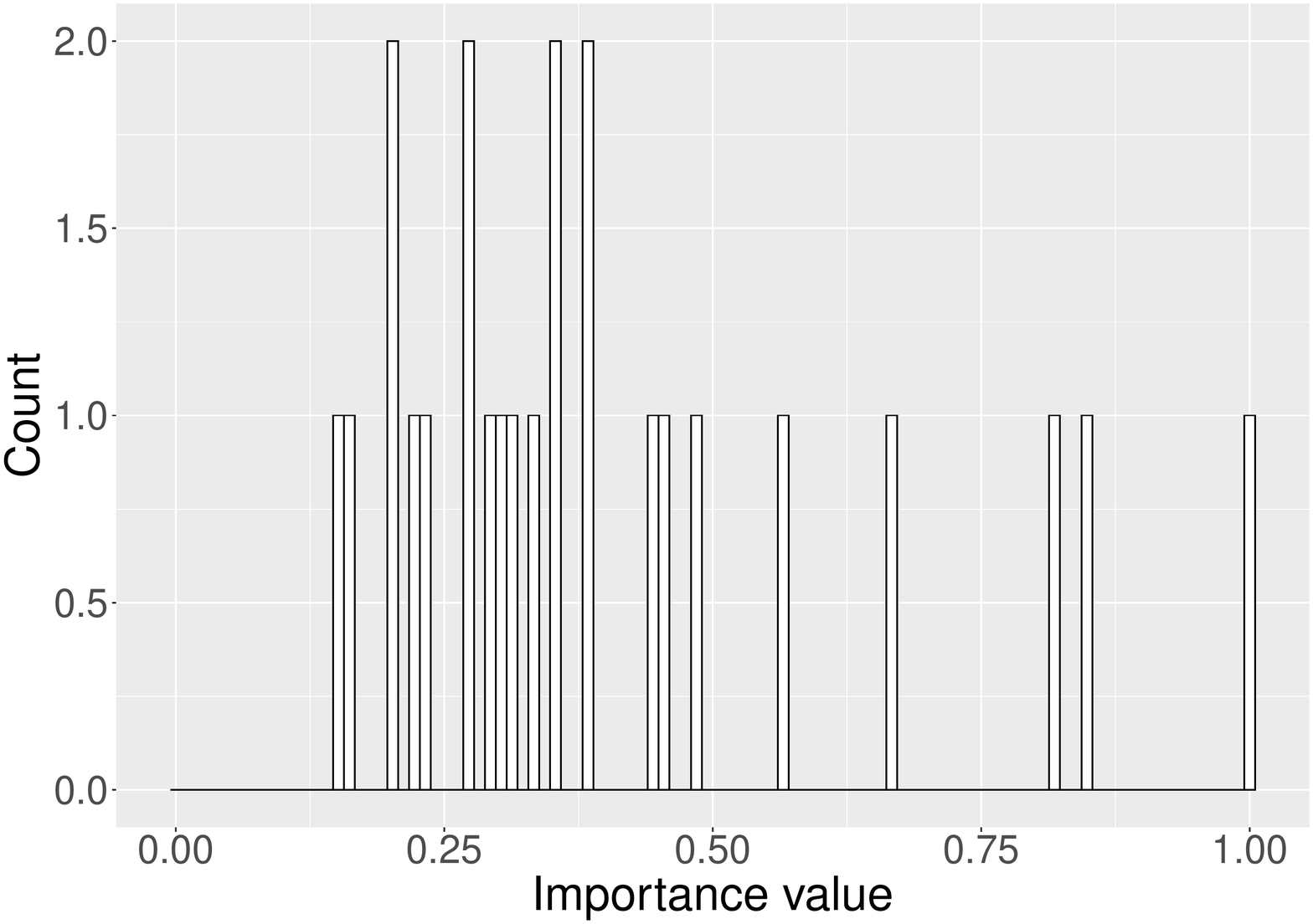}
		\caption{NATOPS dataset.}
		\label{fig:hist:c}
	\end{subfigure}
	\hfill 
	\begin{subfigure}{0.49\textwidth}
		\includegraphics[width=\textwidth]{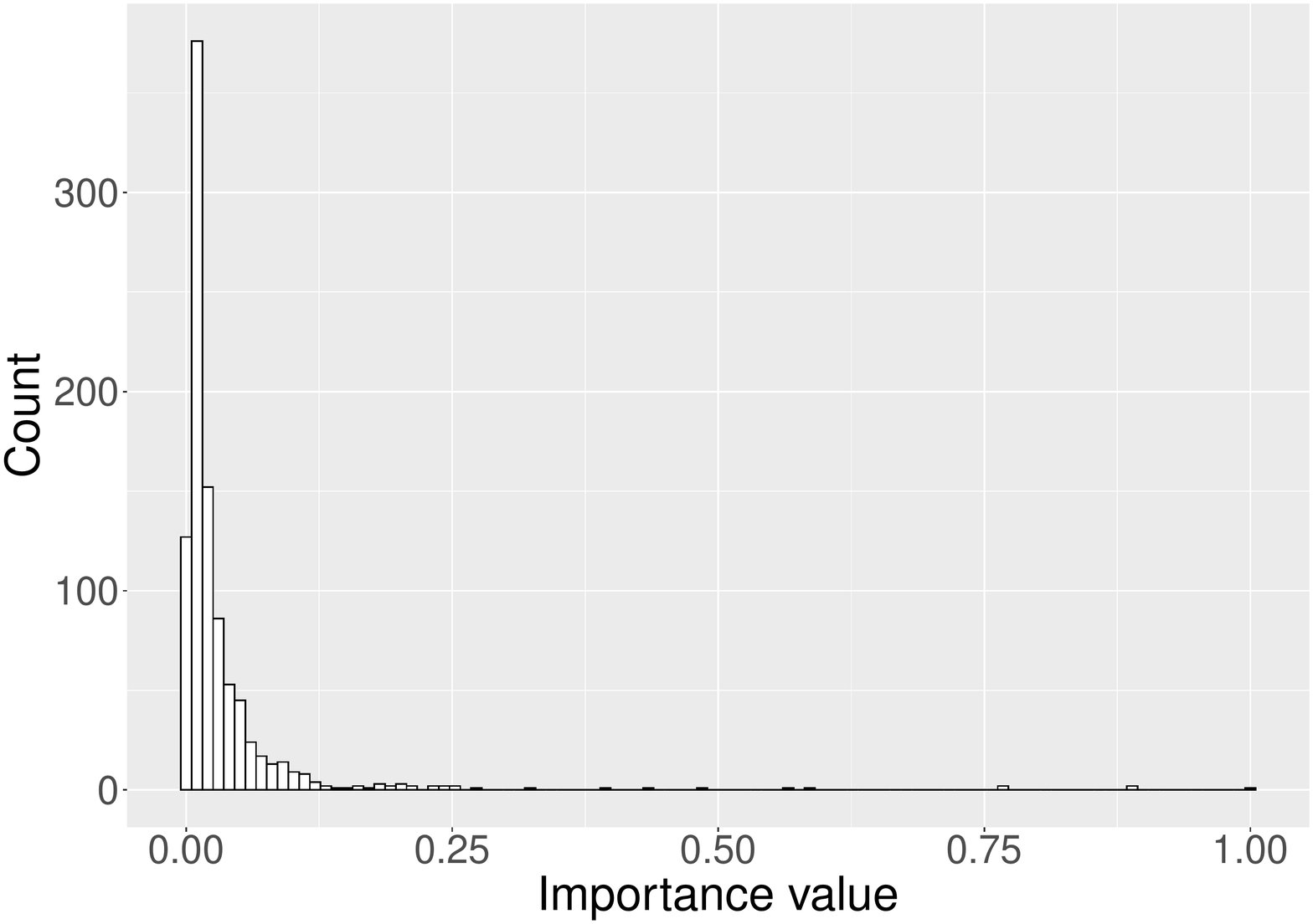}
		\caption{PEMS-SF dataset.}
		\label{fig:hist:d}
	\end{subfigure}
\caption{Histograms of variable importance values.} 
\label{fig:hists}
\end{figure}

If we relate these values to the cases discussed above we see that, for example, in the case of the PhonemeSpectra dataset all 11 variables are of similar importance and CMFMTS+RF was close to the best results obtained, this means that all variables contain information of interest. In LSST we see that although the 6 variables contain information of interest, some of these variables stand out from the rest. To study in a simple way the distribution of the accumulated importance values, we have chosen to include, Figure~\ref{fig:hists}, the histograms of these values for some cases of interest.
For the BasicMotions dataset, in Figure~\ref{fig:hist:a}, we see 2 variables with much higher importance than the remainder together with a third variable that also stands out. These variables are, in decreasing order of importance: 6, 2, 1, 4, 3, and 5. If we compare these values of importance given by RF with the tree C5.0B of classification shown in Figure~\ref{fig:exampleC5:a}, we see how this tree is composed by 3 nodes, 2 of these nodes have features of the variable 6 and the remaining node has a feature of the variable 1, what matches with the variables with more importance given by RF. For the dataset SelfRegulationSCP1, Figure~\ref{fig:hist:b}, we can see similar behavior. We see a variable significantly separated from the rest and the rest of the variables are close to each other, in terms of importance. These variables are, in decreasing order of importance: 1, 2, 3, 4, 5, and 6. If we make the previous comparison with regard to the tree C5.0B of Figure~\ref{fig:exampleC5:b}, we see how the tree is composed of 5 nodes, 4 of these nodes have a feature of the variable 1, more important variable according to the RF. While the remaining node has a feature of variable 2, the second most important variable according to the RF. These examples show a certain relation between the variables with more importance according to the RF and those used by a simple classifier such as C5.0B. In the case of NATOPS dataset where only some of the 24 variables contain the most relevant information. In Figure~\ref{fig:hist:c}, we can see that it is more difficult to obtain well-differentiated groups of variables according to their importance. In this case, we can see that the 3 variables with the greatest importance are significantly distanced from the remainder, with importance values higher than 0.75. Depending on the information sought and the difficulty of the problem, we could decide to lower the threshold, create different groups of variables, etc. These histograms are especially interesting for datasets with a large number of variables. For example, the dataset PEMS-SF, Figure~\ref{fig:hist:d}, where only some of the 963 variables have a high importance, giving residual importance to the rest. In this case, there are only 5 variables with an accumulated importance value higher than 0.75. These variables are, in decreasing order of importance: 212, 187, 55, 172, and 604.

\begin{figure}[!ht]
	\includegraphics[width=\textwidth]{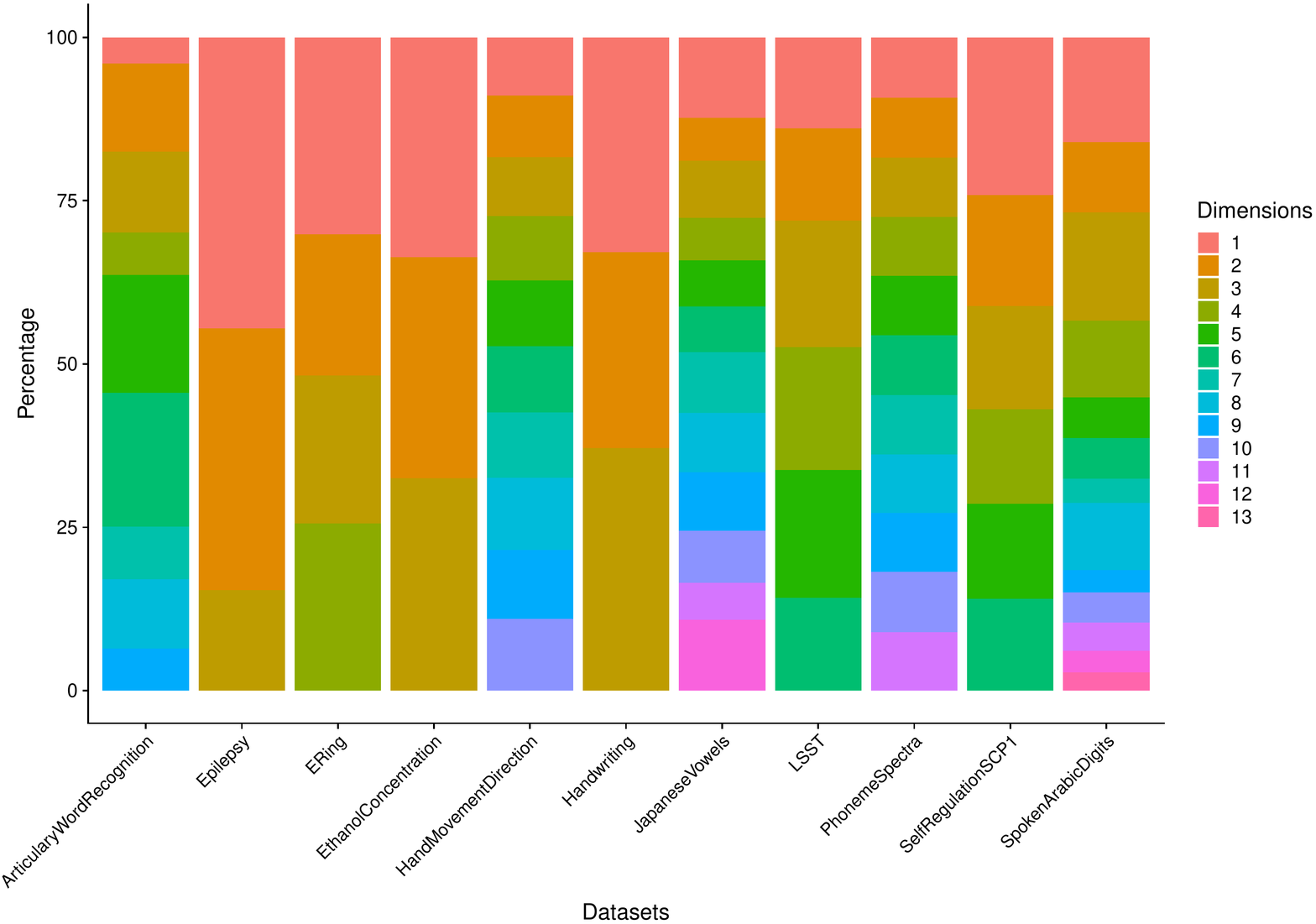}
	\caption{Accumulated importance by variable.}
	\label{fig:impDimension}
\end{figure}

In order to evaluate the relative importance of each variable within a problem, we show graphically the proportion of importance by each variable of the most representative and easily visualized cases in Figure~\ref{fig:impDimension}.

By looking at Figure~\ref{fig:impDimension}, we can determine, for selected cases, the variables that have the greatest potential, so that efforts can be directed at improving the information recorded on those variables or processing them with greater attention. For the ArticularyWordRecognition dataset, we see that variables 1, 4, 7, and 9 are of less importance. ERing datasets shows a great importance accumulated in variables 1 and 4. Epilepsy dataset has much of its useful information in variables 1 and 2. EthanolConcentration and Handwriting show similar behavior. The HandMovementDirection dataset shows how variables 8, 9, and 10 have greater cumulative importance. In the case of SpokenArabicDigits the most important variables (1, 2, 3, 4, and 8) can be clearly differentiated from the rest (5, 6, 7, 9, 10, 11, 12, and 13). For the LSST dataset, we see that variables 3, 4, and 5 have of higher weight. JapaneseVowels dataset shows that variables 8, 9, and 13 have of greater importance. With these results, the preprocessing of the data can be modified in such a way as to improve the recording of the data of these variables or to give them greater importance in the learning process.

\section{Conclusion}
\label{conclusion}
In this paper we have presented a method that allows to apply a feature-based approach designed for univariate time series classification problems to MTSC problems, namely CMFMTS. This method enables the use of traditional classification algorithms on MTSC problems, considerably expanding the tools available to deal with this type of problem.  Furthermore, this allows to achieve interpretable results in a field where the solutions obtained are characterized by their complexity, which is directly related to the number of variables of the problem.

We have published the software of our proposal so it can be used freely. Also, we have published all the information to make our work fully reproducible. Our proposal has been evaluated on 30 datasets from different fields, obtained from the UEA repository. We have focused on tree-based algorithms because of their high interpretability, comparing their results with the main state-of-the-art algorithms of MTSC.

CMFMTS offers very competitive accuracy results in comparison with the main state-of-the-art algorithms. Using the CD diagram, we see that there is no significant statistical difference, for an $\alpha$ of 0.05, between state-of-the-art algorithms and our proposal best case, CMFMTS+RF. We can conclude that these methods work equally well from a statistical point of view.

The interpretability of the results obtained is a significant advantage of our proposal compared to other methods of the state-of-the-art. CMFMTS allows us to relate representative characteristics of the time series with the classification made, depending on the algorithm used for modeling the problem. The trees obtained by the C5.0B algorithm are a clear example of this. Even less interpretable algorithms than the classification trees offer all kinds of valuable information. The use of the RF algorithm and the Mean Decrease Gini Importance as a measure of evaluation of the features used has allowed us to identify which features have a high potential of information in each problem, revealing very different behavior between different datasets. This has allowed to identify the most valuable features for each case, relating each dataset with the behaviors of interest associated with those features.

We have verified the existence of a set of features that maintains high importance throughout different datasets. However, there are also certain cases where other features that are less important on average offer the best results. These features are usually related to characteristic behaviors of the time series. This fact reinforces the idea that the selected features must be supported by a strong theoretical foundation in the field of time series and not be selected only through a purely experimental approach. In addition, the nature of the problem will define which features collect its information in the best possible way.

Finally, based on the accumulated importance per variable of our proposal, we have corroborated that, in a significant part of the processed datasets, not all the variables of an MTS have the same importance when it comes to solving a problem. In a significant part of the processed datasets, we have identified between 1 and 4 variables that contained most of the information. This phenomenon is typical of MTS. These results have a direct impact on future classification processes, which can benefit significantly from the additional information generated.

\section{Acknowledgements}
This research has been partially funded by the following grants: TIN2016-81113-R from the Spanish Ministry of Economy and Competitiveness, and P12-TIC-2958 from Andalusian Regional Government, Spain. Francisco J. Bald\'an holds the FPI grant BES-2017-080137 from the Spanish Ministry of Economy and Competitiveness.

\section{Bibliography}
\bibliography{mvcmfts}

\end{document}